\documentclass{article}

\usepackage{microtype}
\usepackage{graphicx}
\usepackage{subfigure}
\usepackage{booktabs} % for professional tables

\usepackage{hyperref}

\usepackage{threeparttable}
\usepackage{multirow}
\usepackage{makecell}
\usepackage{enumerate}

\usepackage[accepted]{icml2024}

\usepackage{amsmath}
\usepackage{amssymb}
\usepackage{mathtools}
\usepackage{amsthm}

\usepackage[capitalize,noabbrev]{cleveref}

\theoremstyle{plain}

\theoremstyle{definition}

\theoremstyle{remark}

\usepackage[textsize=tiny]{todonotes}

\icmltitlerunning{Diagnosing the Compositional Knowledge of Vision Language Models from a Game-Theoretic View}

\begin{document}

\twocolumn[
\icmltitle{Diagnosing the Compositional Knowledge of Vision Language Models from a Game-Theoretic View}




\begin{icmlauthorlist}
\icmlauthor{Jin Wang}{allaff}
\icmlauthor{Shichao Dong$^{\;\textnormal{4}}$}{}
\icmlauthor{Yapeng Zhu$^{\;\textnormal{4}}$}{}
\icmlauthor{Kelu Yao$^{\;\textnormal{3}}$}{}
\icmlauthor{Weidong Zhao$^{\;\textnormal{3}}$}{}
\icmlauthor{Chao Li$^{\;\textnormal{3}}$}{}
\icmlauthor{Ping Luo$^{\;\textnormal{2\ 1}}$}{}
\end{icmlauthorlist}

\icmlaffiliation{allaff}{Department of Computer Science, The University of Hong Kong, Hong Kong 
$^\textnormal{2}$ Shanghai AI Laboratory, China
$^\textnormal{3}$ Zhejiang Laboratory, Hangzhou, China
$^\textnormal{4}$ Baidu Inc, Beijing, China 
}

\icmlcorrespondingauthor{Ping Luo}{pluo@cs.hku.hk}
\icmlcorrespondingauthor{Chao Li}{lichao@zhejianglab.com}

\icmlkeywords{Machine Learning, ICML}

\vskip 0.3in
]



\printAffiliationsAndNotice{}  % leave blank if no need to mention equal contribution

\begin{abstract}
Compositional reasoning capabilities are usually considered as fundamental skills to characterize human perception.
Recent studies show that current Vision Language Models (VLMs) surprisingly lack sufficient knowledge with respect to such capabilities. 
To this end, we propose to thoroughly diagnose the composition representations encoded by VLMs, systematically revealing the potential cause for this weakness.
Specifically, we propose evaluation methods from a novel game-theoretic view to assess the vulnerability of VLMs on different aspects of compositional understanding, \emph{e.g.}, relations and attributes.
Extensive experimental results demonstrate and validate several insights to understand the incapabilities of VLMs on compositional reasoning, which provide useful and reliable guidance for future studies.
The deliverables will be updated \href{https://vlms-compositionality-gametheory.github.io/}{here}.
\end{abstract}

\section{Introduction}
\label{sec:intro}
%--------- figure 1. --------------
\begin{figure*}[t]
\centering
\includegraphics [width=0.9\textwidth]{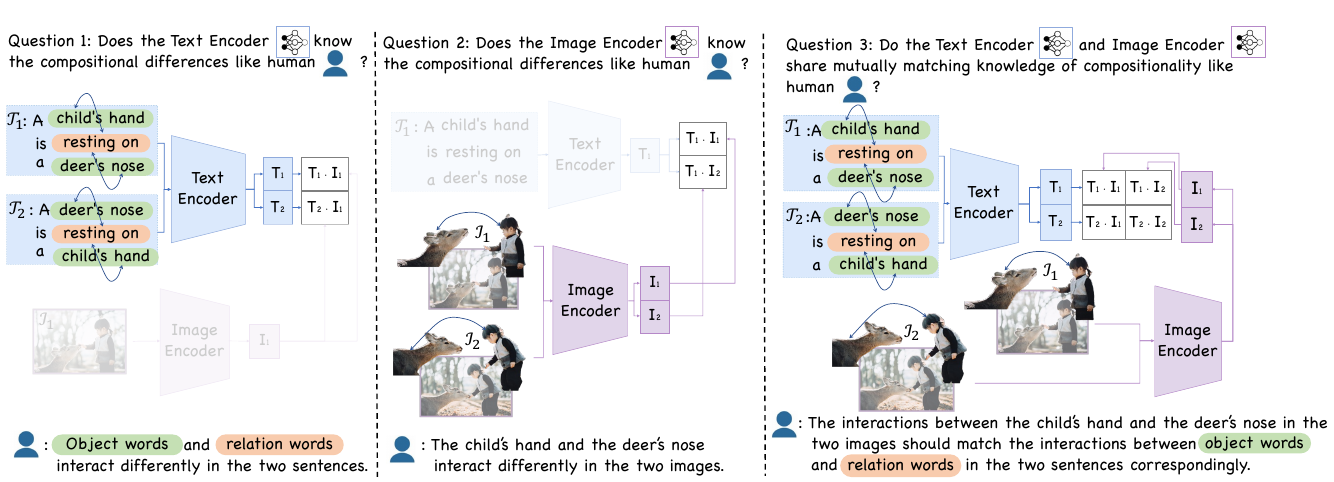}
\caption{
Diagnosing the compositional reasoning capabilities of Vision Language Models (VLMs).
In this paper, we systematically analyze the potential causes for the poor compositional performance of VLMs from each unimodal \textit{separately} and then multimodal \textit{jointly}.
In this way, three insights are obtained and validated correspondingly. Please see Section \ref{sec:intro} for detailed elaborations.
}
\label{fig:1}
\end{figure*}

Recently, Vision Language Models (VLMs) \cite{clip,align,blip,flava,cyclip,filip} have made remarkable strides, which significantly advance a wide range of unimodal and multimodal applications, \emph{e.g.}, object detection \cite{gu2021open,du2022learning,zang2022open}, semantic segmentation \cite{zhou2022extract,li2022languagedriven,xu2022groupvit} and text-to-image generation \cite{SD,dalle-2}.
However, recent studies have unveiled a surprising weakness of state-of-the-art VLMs: they struggle with compositional reasoning capabilities \cite{aro,winoground}, such as object relations and object attributes.
For instance, BLIP \cite{blip} failed to correctly comprehend the subtle differences between ``the horse is eating the grass" and ``the grass is eating the horse" \cite{aro}.
Given the fundamental status of compositionality in human intelligence \cite{cresswell1973logics}, this lack of compositional knowledge has hindered the further development of vision language models.

Previous studies on the compositionality of VLMs mainly focused on two perspectives.
Some studies proposed to evaluate the compositional reasoning capabilities of VLMs in a black-box probing manner \cite{aro,winoground,crepe,sugarcrepe,vlchecklist}.
They usually measured the accuracy performance on whether VLMs correctly retrieved the matching text for a given image between two captions with minimal changes.
Other studies proposed to improve the compositionality of VLMs in an empirical manner by introducing the supervision of scene graphs \cite{herzig2023incorporating,huang2023structure} or curated hard-negative samples \cite{aro,doveh2023teaching}.
However, there still lack in-depth analyses to thoroughly diagnose the internal compositional representations of VLMs, which can help us understand the essential cause of this weakness and provide reliable guidance for future studies.

Therefore, in this paper, we propose to take a further step and conduct detailed analyses on the potential causes of VLMs' poor compositional reasoning capabilities.
Since a VLM usually contains an image encoder and a text encoder as a whole, we propose to comprehensively evaluate VLMs by firstly focusing on the compositional knowledge of each unimodal encoder \textit{separately} and then the multimodal compositional knowledge \textit{jointly}.
Under this scheme, we expect to answer the following questions.

$\bullet$ \textit{Question 1.} \textbf{Does the text encoder of a VLM understand texts compositionally?}

$\bullet$ \textit{Question 2.} \textbf{Does the image encoder of a VLM understand images compositionally?}

$\bullet$ \textit{Question 3.} \textbf{Do the text encoder and the image encoder of a VLM have mutually-matching knowledge on compositionality?}

In this way, such a disentangled representation dissection scheme can help us obtain a more meticulous and fine-grained understanding on VLMs' poor compositional reasoning capabilities.

To answer previous questions, we propose to evaluate the compositional representations inside VLMs from a novel game-theoretic view.
Specifically, we propose several metrics based on the Harsanyi dividend \cite{harsanyi} to assess the sensitivities of VLMs to the changes of different compositionality aspects, \emph{e.g.}, relations and attributes.
The Harsanyi dividend was originally proposed in game theory to measure the interactions between different players, which makes itself a natural metric to dissect the compositional knowledge of DNNs.
Besides, the Harsanyi dividend is related to the Shapley value \cite{shapley-value}, theoretically satisfying \textit{the efficiency, linearity, dummy, symmetry axioms}, which further ensures the trustworthiness of the interpretations for DNNs \cite{ren2022can,ren2023defining,li2023does}.

In this way, we conduct extensive evaluations on five state-of-the-art VLMs \cite{clip,blip,xvlm,aro,flava} with four widely-used datasets \cite{aro,vlchecklist,sugarcrepe,eqben}, obtaining and validating several fine-grained insights on the internal representations of VLMs \emph{w.r.t.} the compositional reasoning capabilities.

$\bullet$ \textit{Insight 1.} \textbf{It is to our surprise that text encoders of VLMs show excellent compositional reasoning capabilities, able to recognize the dominant compositional differences between input texts like human understanding.}

$\bullet$ \textit{Insight 2.}
\textbf{Image encoders of VLMs demonstrate compositional reasoning capabilities to some extent, which are relatively weaker than the corresponding text encoders, partially resulting in the poor compositional performance of VLMs.}

$\bullet$ \textit{Insight 3.} \textbf{Although text encoders and image encoders show certain compositional reasoning capabilities individually, they do not share mutually-matching compositional knowledge, which also partially accounts for the poor compositional abilities of VLMs\footnote{Compared to previous studies \cite{herzig2023incorporating,huang2023structure,doveh2023teaching} which mainly provided intuitive understanding in this aspect, we presented detailed and in-depth analyses from a novel game-theoretic view to further validate this assumption in this paper.}.}

These insights provide a detailed understanding on the potential causes of VLMs' poor compositional knowledge, which can provide beneficial and reliable instructions for future studies.
For instance, to bring more significant performance gain on the compositional reasoning tasks, it may be more effective to design stronger image encoders instead of text encoders for VLMs.

The contributions of our paper are summarized as follows.
1) We conduct a systematical analysis to diagnose the internal representations of VLMs, progressively revealing the potential causes for their weakness in compositional reasoning capabilities.
2) We propose several metrics from a novel game-theoretic view to assess the vulnerability of VLMs on different aspects of compositional understanding, \emph{e.g.}, relations and attributes.
3) Experimental results on various state-of-the-art VLMs and datasets demonstrate and validate several insights on the compositional reasoning capabilities of VLMs, which can help instruct future studies for more effective improvements of VLMs.

\section{Related Work}
\label{sec:RW}
\noindent \textbf{Vision Language Models.}
In recent years, Vision Language Models (VLMs) \cite{2019UNITER,2019VisualBERT,2019LXMERT,2021Align,align,cyclip,flava,liu2023visual,zhu2023minigpt4} have received an increasing research focus, providing benefits for both unimodal and multimodal applications.
Generally, VLMs were trained to generate correspondences between input texts and images.
Their encoded representations were then evaluated on many zero-short or few-short downstream tasks, such as 
CLIP \cite{clip}, X-VLM \cite{xvlm}, BLIP \cite{blip} and etc. 
However, recent studies showed that despite the high performance on dozens of well-established benchmarks \cite{filip, 2021Supervision, 2022PyramidCLIP}, most of the VLMs surprisingly exhibited poor compositional understanding capabilities \cite{aro,winoground}. 

\noindent \textbf{Compositionality of VLMs.} 
Previous studies firstly designed a number of new benchmarks to comprehensively evaluate the compositional reasoning capabilities of VLMs, such as Winoground \cite{winoground}, ARO \cite{aro}, VL-CheckList \cite{vlchecklist}, CREPE \cite{crepe}, SUGARCREPE \cite{sugarcrepe}, COLA \cite{cola}, EQBEN \cite{eqben}, SyViC \cite{cascante2023going} and SPEC \cite{peng2024spec}.
However, these benchmarks usually evaluated the compositionality of VLMs with the accuracy metric, testing whether the paired images and texts could be correctly retrieved among perturbed samples.
Such a black-box probing scheme failed to provide further explanations for the unsatisfying performance.
Besides the development of these benchmarks, other studies focused on improving the compositional performance of VLMs in an empirical manner, such as introducing the guidance of scene graphs \cite{herzig2023incorporating,huang2023structure} and generating curated hard-negative samples \cite{aro,doveh2023teaching,sahin2023enhancing,momeni2023verbs}.
However, there still lacks a thorough representation diagnosis on the compositional reasoning capabilities of VLMs, so as to systematically unveil the essential causes for this weakness.

\noindent \textbf{Interactions of DNNs.}
Considerable studies have focused on quantifying the interactions among input units for diagnosing the representation of DNNs \cite{grabisch1999axiomatic, zhang2020interpreting, zhang2021building, wang2020unified, ren2021unified, wang2021interpreting, yao2023towards, dong2022explaining,chen2023harsanyinet}.
Based on the Shapley value \cite{shapley-value} originally proposed in game theory, some studies \cite{grabisch1999axiomatic,sundararajan2020shapley} proposed interaction metrics to quantify the relationships among the input units, such as the Harsanyi dividend \cite{harsanyi}.
Besides, Zhang \textit{et al.} \yrcite{zhang2021interpreting,zhang2020game} further extended the interaction metric to the multi-order and multivariate interactions, which were applied to explain several phenomena of DNNs \cite{deng2021discovering,wang2020unified,ren2021unified}. 
In comparison, our study aims to provide detailed explanations on the poor compositional reasoning capabilities of VLMs.

\section{Quantifying the compositional knowledge of VLMs with the Harsanyi dividend}
In this paper, we propose several metrics based on the Harsanyi dividend \cite{harsanyi} to evaluate the compositional reasoning capabilities of VLMs from different aspects. 
To this end, we first present a brief introduction to the Harsanyi dividend for better understanding.
\subsection{The Harsanyi dividend}
The Harsanyi dividend was a typical metric in game theory \cite{harsanyi}, which measures the interaction among a set of players. 
Specifically, given a set of players $\mathcal{N} = \{1, 2, ..., n\}$ participating in a game $v$, certain \textit{rewards} can be obtained.
Here, $v(\cdot)$ represents a function to map any subset of players $\mathcal{S}\subseteq \mathcal{N}$ to a real number, representing the obtained numerical \textit{reward}.
Intuitively, during such a game, each player usually does not contribute to the \textit{reward} individually, but interacts with each other, forming different \textit{coalitions}/\textit{patterns} to cause casual effects on the final outcome.
Mathematically, such effects can be measured by the Harsanyi dividend, which is defined as follows:
\begin{equation}
w({\mathcal{S}}|\mathcal{N})=\sum_{\mathcal{S}^{\prime} \subseteq \mathcal{S}}(-1)^{\left|\mathcal{S}^{\prime}\right|-|\mathcal{S}|} \cdot v({\mathcal{S}^{\prime}}).
\end{equation}
Besides, the Harsanyi dividend also satisfies many axioms to theoretically support the fairness and trustworthiness of its calculation \cite{axioms,ren2023defining}. 
Please see Appendix \ref{sec:rationale} for details.

\subsection{Quantifying the compositional knowledge of VLMs}
Based on the definition of the Harsanyi dividend, we then elaborate on how to quantify the compositional knowledge of VLMs with it.
To limit the scope of discussion in this paper, we mainly focus on the aspects of compositionality as follows: relations, objects and attributes. 
To this end, we believe that \textbf{a VLM with a comprehensive and excellent compositional reasoning capability should be sensitive to the changes of objects, relations, attributes, and also, being sensitive to the changes of interactions among them}.
To be specific, let us take the samples in Figure \ref{fig:1} for an example. 

Given the two input captions in Figure \ref{fig:1} showing changes regarding the textual compositionality, they both contain the \textit{same} words but these words have \textit{different} interactions with each other, describing different relations between objects.
To this end, a VLM with an excellent compositional reasoning capability should learn that the object words alone (\emph{i.e.}, \textit{\{child's hand, deer's nose\}}) or the relation words alone (\emph{i.e.}, \textit{\{resting on\}}), have almost the same casual effect on the understanding of each caption.
By contrast, the interactions between the object words and the relation words in each caption should have different casual effects on the understanding of each caption correspondingly.

%--------- figure 1. --------------
\begin{figure}[t]
\centering
\includegraphics [width=0.95\columnwidth]{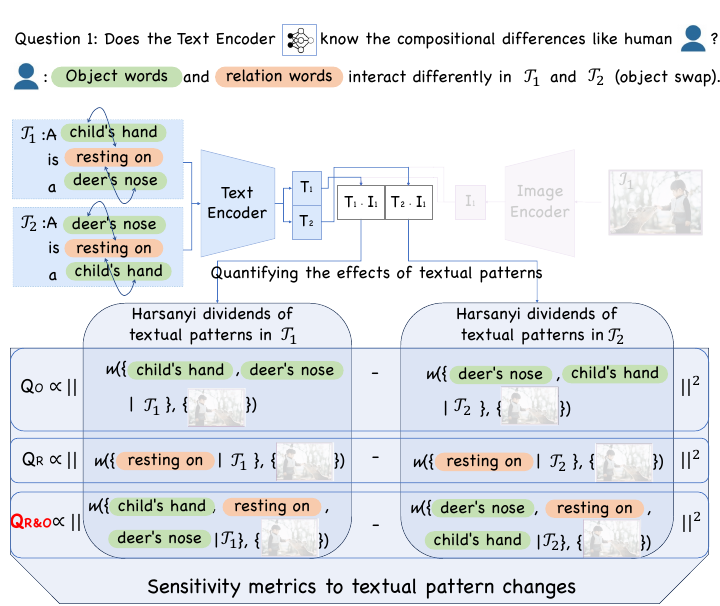}
\caption{
Evaluating the sensitivities of text encoders of VLMs to the changes of textual patterns.
Specifically, given captions $\mathcal{T}_1$ and $\mathcal{T}_2$ with object words being swapped, we design $Q_O$, $Q_R$ and $Q_{R\&O}$ to assess whether text encoders react correctly to the fine-grained changes of compositionality. 
In this case, $Q_{R\&O}$, which measures the interaction changes between object words and relation words, should be of a greater value than $Q_O$ and $Q_R$.
Here $\text{T}_1 \cdot \text{I}_1$ represents the cosine similarity between the normalized text embedding $\text{T}_1$ and normalized image embedding $\text{I}_1$.
}
\label{fig:2}
\end{figure}

Besides, given the two input images in Figure \ref{fig:1} showing changes in terms of the visual compositionality, they both contain objects of the \textit{same} identity (\emph{i.e.}, image regions showing the same child and the same deer in these two images.), but these objects interact \textit{differently} with each other.
In this way, between these two images, a VLM with an excellent compositional reasoning capability should learn that image regions of the same object alone (\emph{i.e.}, image regions of the child or the deer alone) should make similar casual effects on the representation of each image.
On the contrary, the interactions between these two object regions should have different casual effects.

In order to meticulously examine the above fine-grained understanding of compositionality inside VLMs, we propose to exploit the Harsanyi dividend to first quantitatively measure the effect of each visual and textual pattern on the output of VLMs.
In this way, we can further quantitatively evaluate whether VLMs show sharp sensitivities to the changes of these patterns, which relate to different aspects of compositionality.
Specifically, we can analogously consider the inference process of VLMs as a game $v(\cdot, \cdot)$ with two sets of players $\mathcal{N^I} = \{1, 2, ..., n^\mathcal{I} \}$ (\emph{e.g.}, all the visual concepts on an image) and $\mathcal{N^T} = \{1, 2, ..., n^\mathcal{T} \}$ (\emph{e.g.}, all the words in a caption).
Here, $v(\cdot, \cdot)$ represents the output of VLMs, which measures the matching similarity between an input image and an input text, \emph{e.g.}, the cosine similarity between an input image embedding and an input text embedding for CLIP \cite{clip}.
The casual effect of the pattern $\mathcal{S^I} \subseteq \mathcal{N^I}$ and $\mathcal{S^T} \subseteq \mathcal{N^T}$ defined by the Harsanyi dividend is then calculated as,
\begin{equation}
\begin{aligned}
\label{coreeq}
    &w(\{ \mathcal{S^I}, \mathcal{S^T} \}| \{ \mathcal{N^I}, \mathcal{N^T}\}) \\
    =&\sum_{\substack{\mathcal{S^{I}}^{\prime} \subseteq \mathcal{S^I}, \\ \mathcal{S^{T}}^{\prime} \subseteq \mathcal{S^T}}}(-1)^{|\mathcal{S^{I}}^{\prime}| -|\mathcal{S^I}| + |\mathcal{S^{T}}^{\prime}| -|\mathcal{S^T}|} \cdot v({{\mathcal{S^I}^{\prime}}}, {{\mathcal{S^T}^{\prime}}}).
\end{aligned}
\end{equation}
where ${\mathcal{S^I}^{\prime}}$ represents the input image with only the visual concepts in the subset $\mathcal{S^I}^{\prime}$ while masking other visual concepts in $\mathcal{N^I} \backslash {\mathcal{S^I}^{\prime}}$; ${\mathcal{S^T}^{\prime}}$ represents the input caption with only the words in the subset $\mathcal{S^T}^{\prime}$ while masking other words in $\mathcal{N^T} \backslash {\mathcal{S^T}^{\prime}}$\footnote{In this paper, the baseline value for masking out image regions/text tokens was set as zero, following previous studies \cite{ancona2019explaining,wang2021interpreting,zhang2020interpreting,zhang2021interpreting,dong2022explaining}.}.
For simplicity, we denote $w(\{ \mathcal{S^I}, \mathcal{S^T} \}| \{ \mathcal{N^I}, \mathcal{N^T}\})$ as $w(\{ \mathcal{S^I}, \mathcal{S^T} \})$ in the following sections. 

With Equation \ref{coreeq} measuring the effects of visual/textual patterns, we then design metrics to evaluate VLMs' sensitivities to the changes of these patterns, so as to fully examine the compositional reasoning capabilities of VLMs in a fine-grained manner.
In the following sections, we start from each unimodal representations of VLMs \textit{separately}, and then to the multimodal representations of VLMs \textit{jointly}.

\section{Can text encoders of VLMs understand texts compositionally?}
\label{sec:text_encoder}
To fully examine the compositional reasoning capabilities of VLMs in a fine-grained manner, we first explore whether text encoders of VLMs encode reliable compositional knowledge in the first place.
Specifically, we propose several metrics based on the Harsanyi dividend to quantitatively evaluate the sensitivities of text encoders to the changes of textual patterns, which relate to different aspects of compositionality.

\begin{figure*}[t]
\centering
{\includegraphics[width=1.0\textwidth]{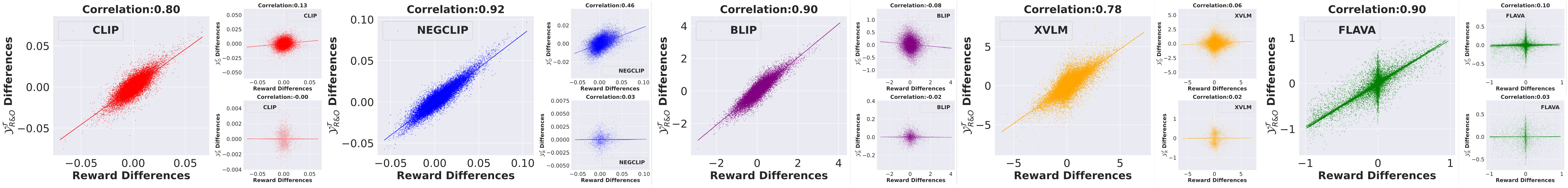}}
\caption{The Pearson correlation coefficients $\rho(\mathcal{X^T}, \mathcal{Y^T})$ between the reward differences $\mathcal{X^T}$ and interaction effect differences on the Visual Genome Relation dataset, \emph{i.e.}, $\mathcal{Y}^\mathcal{T}_R$, $\mathcal{Y}^\mathcal{T}_O$ and $\mathcal{Y}^\mathcal{T}_{R \& O}$.
Each point represents a data sample containing two captions and one image.
Results show that the reward differences between the two captions were mainly related to the interaction changes of object words and relation words, demonstrating that text encoders of VLMs reacted correctly to the textual compositional differences in this dataset.}
\label{fig:relation}
\end{figure*}

As shown in Figure \ref{fig:2}, given an image-text pair $\{\mathcal{I}_1$, $\mathcal{T}_1\}$ and a perturbed text $ \mathcal{T}_2$, which is generated from swapping object words in $\mathcal{T}_1$, VLMs like CLIP \cite{clip} would output two matching scores $v(\mathcal{N}^{\mathcal{I}_1}, \mathcal{N}^{\mathcal{T}_1}))=\text{I}_1\cdot\text{T}_1$ and $v(\mathcal{N}^{\mathcal{I}_1}, \mathcal{N}^{\mathcal{T}_2}))=\text{I}_1\cdot\text{T}_2$, where $\text{I}_1$ represents the normalized image embedding of $\mathcal{I}_1$ and $\text{T}_{1/2}$ represents the normalized text embedding of $\mathcal{T}_{1/2}$.
In this way, the variations between $\text{I}_1\cdot\text{T}_1$ and $\text{I}_1\cdot\text{T}_2$ can be considered as the \textit{coarse} measurement on the sensitivities of text encoders to the changes of a mixture of textual patterns.
However, it still remains uncertain whether text encoders of VLMs react to each specific texture pattern in a correct manner individually.
To this end, we expect to take a step further and comprehensively analyze \textit{fine-grained} sensitivities of text encoders, regarding different aspects of compositionality.
Mathematically, we propose the sensitivity metric as follows,
\begin{equation}
\label{eq:Q}
\begin{aligned}
& {Q}(\mathcal{T}_1, \mathcal{T}_2, \mathcal{I}_1)\\
= & \frac{1}{\mathcal{Z}_{sens}^\mathcal{T}}||w(\{\mathcal{N}^{\mathcal{I}_1}, {\mathcal{S}^{\mathcal{T}_1}}\}) - w(\{\mathcal{N}^{\mathcal{I}_1}, {\mathcal{S}^{\mathcal{T}_2}}\})||^2
\end{aligned}
\end{equation}
where $\mathcal{Z}_{sens}^\mathcal{T} = E_{\mathcal{T}^\prime \in \{\mathcal{T}_1,\mathcal{T}_2\}}E_{\mathcal{{S^T}^\prime}}||w(\{\mathcal{N}^{\mathcal{I}_1}, {\mathcal{S}^{\mathcal{T}^\prime}}\})||^2$ is used for normalization.
Intuitively, this metric measures detailed textual casual effect changes between text $\mathcal{T}_1$ and $\mathcal{T}_2$ when given the image $\mathcal{I}_1$\footnote{Note that to control the input variables for better clarity, in Eq. \ref{eq:Q}, we do not mask out any image regions when only analysing text pattern casual effect changes.
}.
In this way, given different subsets of words $\mathcal{S}^{\mathcal{T}_{1}}$ and  $\mathcal{S}^{\mathcal{T}_{2}}$, we mainly compute five types of sensitivity metrics in implementations.

$\bullet$ \textbf{The sensitivity metric of relation words} $Q_R$. 
This metric measures the casual effect changes of only relation words on the output of text encoders of VLMs, where $\mathcal{S}^{\mathcal{T}_{1}}$ and $\mathcal{S}^{\mathcal{T}_{2}}$ contain relation words only.

$\bullet$ \textbf{The sensitivity metric of attribute words} $Q_A$.
This metric measures the casual effect changes of only attribute words on the output of text encoders of VLMs, where $\mathcal{S}^{\mathcal{T}_{1}}$ and $\mathcal{S}^{\mathcal{T}_{2}}$ contain attribute words only.

$\bullet$ \textbf{The sensitivity metric of object words} $Q_O$.
This metric measures the casual effect changes of only object words on the output of text encoders of VLMs, where $\mathcal{S}^{\mathcal{T}_{1}}$ and $\mathcal{S}^{\mathcal{T}_{2}}$ contain object words only.

$\bullet$ \textbf{The sensitivity metric of interaction between relation words and object words} $Q_{R\&O}$. 
This metric measures the casual effect changes of interactions between relation words and object words, where $\mathcal{S}^{\mathcal{T}_{1}}$ and $\mathcal{S}^{\mathcal{T}_{2}}$ contain both relation words and object words.

$\bullet$ \textbf{The sensitivity metric of interaction between attribute words and object words} $Q_{A\&O}$. 
This metric measures the casual effect changes of interactions between attribute words and object words, where $\mathcal{S}^{\mathcal{T}_{1}}$ and $\mathcal{S}^{\mathcal{T}_{2}}$ contain both attribute words and object words.

\textbf{Experiment protocols.} Based on the metrics, we systematically analyzed the compositional knowledge of text encoders of various VLMs. 
Specifically, we evaluated text encoders of five state-of-the-art VLMs: CLIP \cite{clip}, BLIP \cite{blip}, NEGCLIP \cite{aro}, XVLM \cite{xvlm} and FLAVA \cite{flava}.
The evaluations were conducted on three popular benchmarks: ARO \cite{aro}, SUGARCREPE \cite{sugarcrepe} and VL-CheckList \cite{vlchecklist}. 
Due to the page limitation, please see Appendix \ref{supp:text_side} for results on SUGARCREPE and VL-CheckList.

\begin{table}[t]
\caption{Evaluating the compositional sensitivities of text encoders of VLMs.
In the Visual Genome Relation dataset, the object words are swapped to obtain perturbed texts. 
Results show that $Q_{R\&O}$ (bold) have larger values than $Q_{R}$ and $Q_{O}$ in this dataset across different VLMs, demonstrating that text encoders of various VLMs exhibit accurate sensitivities to the changes of textual patterns.}
\label{aro-relation}
\begin{center}
{\linespread{1.0}
\setlength\tabcolsep{6pt}
\scriptsize
\begin{threeparttable}
\begin{tabular}{l l c c c}
\hline
{Dataset} &{Models}
& $Q_O$ & $Q_R$ & $Q_{R \& O}$ \\
\hline
\hline
\multirow{5}{*}{Visual Genome Relation}&CLIP  &4.5e-3 & 9.8e-6 & \textbf{1.3e-2}  \\
&NEGCLIP & 3.7e-3 & 1.1e-5 & \textbf{2.0e-2}  \\
&BLIP & 3.3e-2 & 1.3e-4 & \textbf{1.7e-1} \\
&XVLM & 5.0e-2& 1.1e-3 & \textbf{1.4e-1} \\
&FLAVA & 7.4e-2& 3.0e-2 & \textbf{4.3e-1} \\
\hline
\end{tabular}
\end{threeparttable}
}
\end{center}

\end{table}

%--------- figure 1. --------------
\begin{figure}[t]
\centering
\includegraphics [width=0.9\columnwidth]{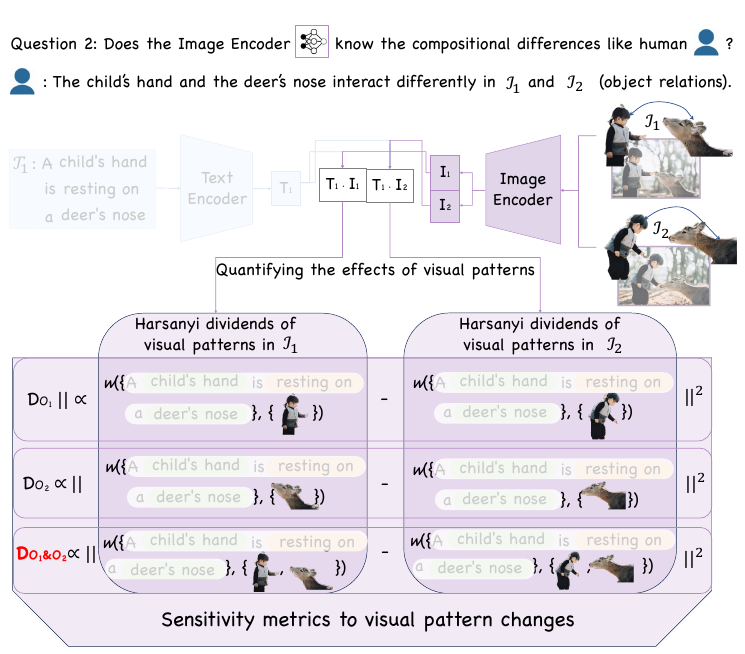}
\caption{
Evaluating the sensitivities of image encoders of VLMs to the changes of visual patterns.
Specifically, given images $\mathcal{I}_1$ and $\mathcal{I}_2$ with object relations being altered, we design $D_{O_1}$, $D_{O_2}$ and $D_{O_1\&O_2}$ to assess whether image encoders react correctly to the fine-grained changes of visual compositionality. 
In this case, $D_{O_1\&O_2}$, which measures the relation changes between objects, should be of a greater value than $D_{O_1}$ and $D_{O_2}$.
}
\label{fig:3}
\end{figure}

\noindent \textbf{Experimental results on ARO benchmark.} 
In this experiment, we mainly focused on the attribute and relation aspects of compositionality, using two of the sub-datasets in the ARO benchmark: Visual Genome Attribution and Visual Genome Relation \cite{visualgenome,gqa}.
Each sample in the dataset includes one image and two captions with minimal differences.
We here present results on the Visual Genome Relation dataset in the main paper.
Please see Appendix \ref{supp:text_side} for results on the Visual Genome Attribution dataset.

In the Visual Genome Relation dataset, the input correct and wrong captions are in the templates of ``\textit{[object 1] [relation] [object 2]}" and ``\textit{[object 2] [relation] [object 1]}" with the \textit{SWAP} manipulation. 
To this end, we mainly evaluated the sensitivity metrics of relation words, object words and the interactions between relation words and object words, \emph{i.e.}, $Q_R$, $Q_O$ and $Q_{R \& O}$. 
Masks denoting the relation words and object words are obtained directly based on the templates.
Based on human understanding, the compositional differences between the correct and wrong captions in this dataset should be largely reflected by the interaction changes between relation words and object words (\emph{i.e.}, $Q_{R \& O}$).
By contrast, the casual effect changes of relation words alone and object words alone should be less significant. 
Results are reported in Table \ref{aro-relation}, where $Q_{R \& O}$ has the highest values among metrics for various VLMs.
\textbf{It is to our surprise that despite the poor performance of VLMs on this benchmark, text encoders of VLMs did recognize the dominant compositional differences between captions in the relation-object aspect, similar to human understanding}.

To further examine the compositional knowledge of text encoders, we propose to calculate the Pearson correlation coefficients between the reward differences and interaction effect differences. 
Specifically, we calculated the reward differences $\mathcal{X^T}$ and interaction effect differences $\mathcal{Y^T}$ as follows: $\mathcal{X^T} = v(\mathcal{N}^{\mathcal{I}_1}, \mathcal{N}^{\mathcal{T}_1}) - v(\mathcal{N}^{\mathcal{I}_1}, \mathcal{N}^{\mathcal{T}_2})$; $\mathcal{Y^T} = w(\mathcal{N}^{\mathcal{I}_1}, \mathcal{S}^{\mathcal{T}_1}) - w(\mathcal{N}^{\mathcal{I}_1}, \mathcal{S}^{\mathcal{T}_2})$.
We then calculated the Pearson correlation coefficients $\rho (\mathcal{X^T}, \mathcal{Y^T})$ for different subsets of words $\mathcal{S}^{\mathcal{T}_{1}}$ and  $\mathcal{S}^{\mathcal{T}_{2}}$, \emph{i.e.}, $\rho (\mathcal{X^T}, \mathcal{Y}^\mathcal{T}_{O})$, $\rho (\mathcal{X^T}, \mathcal{Y}^\mathcal{T}_{R})$, $\rho (\mathcal{X^T}, \mathcal{Y}^\mathcal{T}_{R\&O})$.
Here $\mathcal{Y}^\mathcal{T}_{O}$ represents the casual effect changes when $\mathcal{S}^{\mathcal{T}_{1}}$ and  $\mathcal{S}^{\mathcal{T}_{2}}$ only contain object words.
$\mathcal{Y}^\mathcal{T}_{R}$ represents the casual effect changes when $\mathcal{S}^{\mathcal{T}_{1}}$ and  $\mathcal{S}^{\mathcal{T}_{2}}$ only contain relation words.
$\mathcal{Y}^\mathcal{T}_{R\&O}$ represents the casual effect changes when $\mathcal{S}^{\mathcal{T}_{1}}$ and  $\mathcal{S}^{\mathcal{T}_{2}}$ contain both object and relation words.
As shown in Figure \ref{fig:relation}, $\mathcal{Y}^\mathcal{T}_{R \& O}$ has a significant positive correlation with the final reward differences $\mathcal{X^T}$ among various VLMs.
Such results demonstrate that the reward differences between correct captions and negative captions were mainly caused by the interaction effect changes between relation words and object words, which was surprisingly consistent with human understanding, despite the unsatisfying performance on this benchmark \cite{aro}.

\begin{table}[t]
\caption{Evaluating the compositional sensitivities of image encoders of VLMs with the EQBEN dataset, where object relations alter but each object maintains the same identity within image pairs. 
The maximum metrics are shown in bold.
}
\begin{center}
{\linespread{1.0}
\setlength\tabcolsep{3pt}
\scriptsize
\begin{threeparttable}
\begin{tabular}{c|c | c c c |c c c }
\hline
{Dataset} &{Models}
& $D_{O_1}$ & $D_{O_2}$ & $D_{O_1 \& O_2}$
& $\rho_{O_1}$ & $\rho_{O_2}$ & $\rho_{O_1\&O_2}$\\
\hline
\hline
\multirow{5}{*}{EQBEN}&CLIP  & 1.7e-2 & \textbf{3.1e-2}  &2.0e-2& 0.34&\textbf{0.56}&0.12  \\
    & NEGCLIP &3.7e-2&\textbf{7.5e-2} &3.4e-2 &0.37&\textbf{0.67}&0.05  \\
    & BLIP &5.6e-1&\textbf{7.4e-1} &4.4e-1 &0.36&\textbf{0.46}&0.23 \\
    & XVLM &8.2e-1&\textbf{1.2e0} &8.1e-1 &0.39&\textbf{0.51}&0.13 \\
    & FLAVA &6.9e-1&1.0e0 &\textbf{1.2e0} &0.20&\textbf{0.33}&0.20 \\
\hline
\end{tabular}
\end{threeparttable}
}
\end{center}
\label{eqben-relation}
\end{table}

\section{Can image encoders of VLMs understand images compositionally?}
In previous analyses, we surprisingly find text encoders of VLMs demonstrate sharp sensitivities to different aspects of compositionality in texts. 
Such experimental results significantly reduce the accountability of text encoders for the poor compositionality performance of VLMs.
To rule out other potential causes, we then turn our concentration on exploring whether image encoders should be held accountable for this weakness.

Similar to the analyses on text encoders of VLMs, we hope to answer the following key question in this section: do image encoders correctly know the fine-grained compositional differences between two images?
As shown in Figure \ref{fig:3}, given an image-text pair $\{\mathcal{I}_1, \mathcal{T}_1\}$ and a perturbed image $\mathcal{I}_2$, where the relation between the child and the deer varies.
It is expected that the relation changes between the child and the deer should be considered the essential compositional differences between images $\mathcal{I}_1$ and  $\mathcal{I}_2$. 
In the meantime, casual effect changes of the child alone or the deer alone should be considered less significant.
To diagnose such detailed visual compositional understanding of VLMs, we propose to calculate the following sensitivity metric,
\begin{equation}
\begin{aligned}
& {D}(\mathcal{I}_1, \mathcal{I}_2, \mathcal{T}_1)\\
= & \frac{1}{\mathcal{Z}_{sens}^\mathcal{I}}||w(\{\mathcal{S}^{\mathcal{I}_1}, {\mathcal{N}^{\mathcal{T}_1}}\}) - w(\{\mathcal{S}^{\mathcal{I}_2}, {\mathcal{N}^{\mathcal{T}_1}}\})||^2
\end{aligned}
\label{eq:I}
\end{equation}
where $\mathcal{Z}_{sens}^\mathcal{I} = E_{\mathcal{I}^\prime \in \{\mathcal{I}_1,\mathcal{I}_2\}}E_{\mathcal{{S^I}^\prime}}||w(\{\mathcal{S}^{\mathcal{I}^\prime}, {\mathcal{N}^{\mathcal{T}_1}}\})||^2$ is calculated for normalization.
Intuitively, this metric measures detailed visual casual effect changes between $\mathcal{I}_1$ and $\mathcal{I}_2$ when given the text $\mathcal{T}_1$\footnote{Similar to Eq. \ref{eq:Q}, we do not mask out any text tokens when only analysing visual pattern casual effect changes for clarity.}.
To limit the discussion in this paper, we only focus on image pairs $\mathcal{I}_1$ and  $\mathcal{I}_2$, which contain the same object pair but show different object relations as in Figure \ref{fig:3}. 
We leave analyses on more complex visual cases for future studies.
To this end, we mainly calculated two types of sensitivity metrics given different sets of visual concepts $\mathcal{S}^{\mathcal{I}_1}$ and $\mathcal{S}^{\mathcal{I}_2}$ in implementations.

$\bullet$ \textbf{The sensitivity metric of each object} $D_{O_1}/D_{O_2}$. This metric measures the casual effect changes of each individual object region on the output of image encoders of VLMs, where $\mathcal{S}^{\mathcal{I}_{1}}$ and $\mathcal{S}^{\mathcal{I}_{2}}$ contain the object regions of the same identity, \emph{i.e.}, object $O_{1}/O_{2}$.

$\bullet$ \textbf{The sensitivity metric of relation within a pair of objects} $D_{O_{1}\&O_{2}}$. This metric measures the casual effect change of the relation within a pair of objects $O_{1}$ and $O_{2}$, where $\mathcal{S}^{\mathcal{I}_{1}}$ and $\mathcal{S}^{\mathcal{I}_{2}}$ contain both object regions.

\noindent \textbf{Experiment protocols.} Following previous analyses, we continued to evaluate the compositional knowledge of image encoders for CLIP, NEGCLIP, BLIP, XVLM and FLAVA.
As for benchmarks, we exploited the EQBEN dataset \cite{eqben} collected from natural videos \cite{zhou2018towards,ji2020action,wang2022geb+} or synthetic engines \cite{SD, hertz2022prompt, greff2022kubric}. 
In this dataset, each sample contains two image-text matching pairs sharing minimal differences.
However, to conduct quantitative evaluations with Eq. \ref{eq:I}, we need to obtain the pixel-wise mask for each described object on images, which is not provided in the original dataset.
To this end, we carefully selected a subset of data samples (290 image-text pairs) from the original dataset and utilized SAM \cite{SAM} to help obtain the final mask for each described object in images. 
Please see Appendix \ref{supp:eqben:vis} for annotations\footnote{New annotations are available \href{https://drive.google.com/drive/folders/172_qzVXNrH05EVJmRfGfoV8RbV761MnJ?usp=sharin}{here}.}.

\noindent \textbf{Experimental results on EQBEN benchmark.} 
Besides the proposed metrics,  we also calculated the Pearson correlation coefficients $\rho_{O_1}$, $\rho_{O_2}$ and $\rho_{O_1\&O_2}$ for further evaluations, following Sec. \ref{sec:text_encoder}.
Here $\rho_{O_1/O_2} = \rho (\mathcal{X^I}, \mathcal{Y}^\mathcal{I}_{O_1/O_2})$ and $\rho_{O_1\&O_2} = \rho (\mathcal{X^I}, \mathcal{Y}^\mathcal{I}_{O_1\&O_2})$, where  $\mathcal{X^I} = v(\mathcal{N}^{\mathcal{I}_1}, \mathcal{N}^{\mathcal{T}_1}) - v(\mathcal{N}^{\mathcal{I}_2}, \mathcal{N}^{\mathcal{T}_1})$; $\mathcal{Y^I} = w(\mathcal{N}^{\mathcal{I}_1}, \mathcal{S}^{\mathcal{T}_1}) - w(\mathcal{N}^{\mathcal{I}_2}, \mathcal{S}^{\mathcal{T}_1})$.
$\mathcal{Y}^\mathcal{I}_{O_1/O_2}$ represents the casual effect changes when $\mathcal{S}^{\mathcal{I}_{1}}$ and  $\mathcal{S}^{\mathcal{I}_{2}}$ only contain the same object $O_1$/$O_2$. 
$\mathcal{Y}^\mathcal{I}_{O_1\&O_2}$ represents the casual effect changes when $\mathcal{S}^{\mathcal{I}_{1}}$ and  $\mathcal{S}^{\mathcal{I}_{2}}$ contain both objects $O_1$ and $O_2$. 

Results are summarized in Table \ref{eqben-relation}. On the one hand, different from the trend in Table \ref{aro-relation} where text encoders show sharp sensitivities to the relation changes between objects (\emph{i.e.}, $Q_{R\&O}$ being larger than $Q_{O}$ and $Q_{R}$.), $D_{O_1 \& O_2}$ had a smaller value than $D_{O_1}$ and $D_{O_2}$ for most VLMs, which shows that image encoders demonstrated less accurate compositional sensitivities regarding the changes of object relations.
They were more sensitive to the mild and less significant changes of each object alone, instead of the major relation changes between objects.
Besides, the coefficients results show that the reward differences within the image pairs were not mostly related to the relation changes between objects, \emph{i.e.}, $\rho_{O_1\&O_2}$ did not have the largest value than $\rho_{O_1}$ and $\rho_{O_2}$, which was less consistent with human understanding compared to text encoders of VLMs.
In summary, the above results show that \textbf{image encoders demonstrated weaker compositional reasoning capabilities, which may partially result in the overall poor compositional performance of VLMs.}

\section{Do text encoders and image encoders have matching compositional knowledge?}
Based on previous analyses, we find that text encoders and image encoders of VLMs both demonstrate certain sensitivities to the compositional changes of input, though image encoders are less sensitive.
To further comprehensively examine the compositional knowledge of VLMs, we then evaluate whether text encoders and image encoders have mutually matching compositional knowledge.
In other words, do image encoders correctly consider the interaction between object words and relation words in texts as the relations between objects in images, or mistakenly consider as each object alone?
Similarly, do text encoders correctly consider the relations between objects in images as the interactions between object words and relation words in texts, or mistakenly consider as the object/relation words alone?

To evaluate the correspondence between the compositional knowledge encoded inside text encoders and image encoders, we propose to compute the following modified metrics for image-text pairs describing relations between two objects,
\begin{equation}
\begin{aligned}
& {Q_{\mathcal{T}:R\&O \xrightarrow{}\mathcal{I}:(\cdot)}} \\
=& \frac{1}{\mathcal{\hat{Z}}_{sens}^\mathcal{T}}{||w(\{\mathcal{S}^{\mathcal{I}_1}, {\mathcal{S}^{\mathcal{T}_1}_{R\&O}}\}) - w(\{\mathcal{S}^{\mathcal{I}_1}, {\mathcal{S}^{\mathcal{T}_2}_{R\&O}}\})||^2}
\end{aligned}
\label{eq-dt2i}
\end{equation}
\begin{equation}
\begin{aligned}
& {D_{\mathcal{I}:O_1\&O_2 \xrightarrow{}\mathcal{T}:(\cdot)}} \\
=& \frac{1}{\mathcal{\hat{Z}}_{sens}^\mathcal{I}}{||w(\{\mathcal{S}^{\mathcal{I}_1}_{O_1\&O_2}, {\mathcal{S}^{\mathcal{T}_1}}\}) - w(\{\mathcal{S}^{\mathcal{I}_2}_{O_1\&O_2}, {\mathcal{S}^{\mathcal{T}_1}}\})||^2}
\end{aligned}
\label{eq-di2t}
\end{equation}
where $\mathcal{\hat{Z}}_{sens}^\mathcal{T} = E_{\mathcal{T}^\prime \in \{\mathcal{T}_1,\mathcal{T}_2\}}E_{\mathcal{{S^T}^\prime}}E_{\mathcal{{S}}^{\mathcal{I}_1}}||w(\{\mathcal{S}^{\mathcal{I}_1}, {\mathcal{S}^{\mathcal{T}^\prime}_{R\&O}}\})||^2$ and  $\mathcal{\hat{Z}}_{sens}^\mathcal{I} = E_{\mathcal{I}^\prime \in \{\mathcal{I}_1,\mathcal{I}_2\}}E_{\mathcal{{S^I}^\prime}}E_{\mathcal{S}^{\mathcal{T}_1}}||w(\{\mathcal{S}^{\mathcal{I}^\prime}_{O_1\&O_2}, {\mathcal{S}^{\mathcal{T}_1}}\})||^2$  are used for normalization.
Here, for $\mathcal{T}^\prime \in \{\mathcal{T}_1,\mathcal{T}_2\}$, $\mathcal{S}^{\mathcal{T}^\prime}_{R\&O}$ denotes the object words and relation words in texts; for $\mathcal{I}^\prime \in \{\mathcal{I}_1,\mathcal{I}_2\}$, $\mathcal{S}^{\mathcal{I}^\prime}_{O_1\&O_2}$ denotes the image regions of object $O_1$ and object $O_2$. 

\begin{figure}[t]
\centering
\includegraphics [width=1.0\columnwidth]{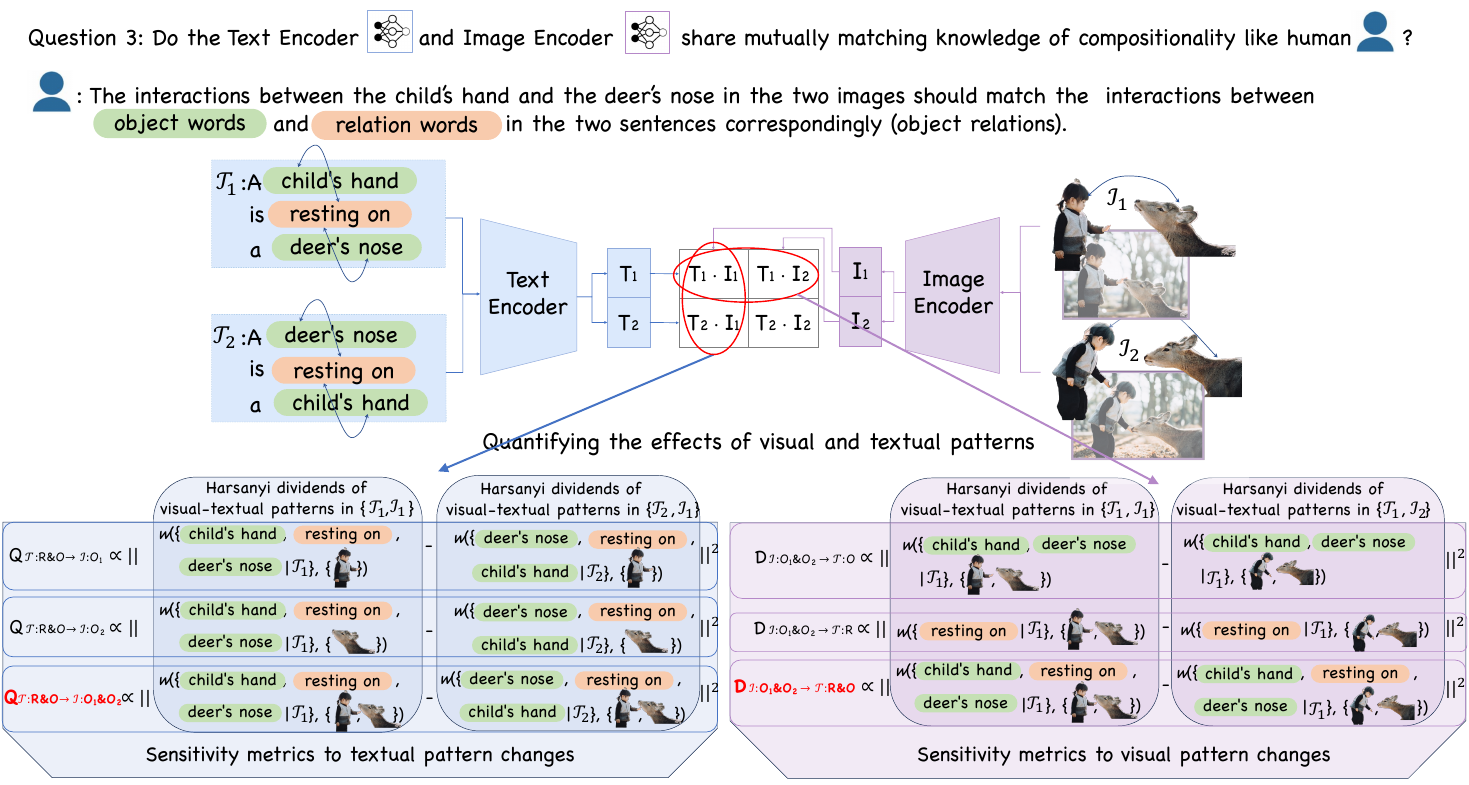}
\caption{
Evaluating whether image encoders and text encoders of VLMs possess mutually matching compositional knowledge with modified sensitivity metrics.
Specifically, given image-text pairs $\{\mathcal{I}_1,\mathcal{T}_1\}$ and $\{\mathcal{I}_2, \mathcal{T}_2\}$ sharing minimal differences of object relations, we design ${Q_{\mathcal{T}:R\&O \xrightarrow{}\mathcal{I}:O_1}}$, ${Q_{\mathcal{T}:R\&O \xrightarrow{}\mathcal{I}:O_2}}$ and ${Q_{\mathcal{T}:R\&O \xrightarrow{}\mathcal{I}:O_1\& O_2}}$ to assess whether image encoders obtain the corresponding compositional knowledge for text encoders. 
Besides, we also design $D_{\mathcal{I}:O_1\&O_2 \xrightarrow{}\mathcal{T}:O}$, $D_{\mathcal{I}:O_1\&O_2 \xrightarrow{}\mathcal{T}:R}$ and $D_{\mathcal{I}:O_1\&O_2 \xrightarrow{}\mathcal{T}:R\&O}$ to assess whether text encoders obtain the corresponding compositional knowledge for image encoders. 
Please zoom in for better visualization.
}
\label{fig:4}
\end{figure}

Intuitively, Eq. \ref{eq-dt2i} aims to examine VLMs by measuring what components of images are more related to the interactions between object words and relation words.
As shown in Figure \ref{fig:4}, we mainly calculated ${Q_{\mathcal{T}:R\&O \xrightarrow{}\mathcal{I}:O_1/O_2}}$ and ${Q_{\mathcal{T}:R\&O \xrightarrow{}\mathcal{I}:O_1\& O_2}}$.
Taking ${Q_{\mathcal{T}:R\&O \xrightarrow{}\mathcal{I}:O_1}}$ as an example, it measures how the interaction changes between object words and relation words within $\mathcal{T}_1$ and $\mathcal{T}_2$ would affect the output of VLMs when only given the image region of the child in $\mathcal{I}_1$. 
In this case, ${Q_{\mathcal{T}:R\&O \xrightarrow{}\mathcal{I}:O_1\& O_2}}$ should be of the greatest value, showing that only when the visual pattern demonstrates the interactions between objects, swapping object words to change their relations in the textual pattern would cause significant influence on the output of VLMs.

%--------- figure 1. --------------

\begin{table}[t]
\caption{Evaluating whether image encoders and text encoders of VLMs possess mutually matching compositional knowledge with the EQBEN dataset, where each sample contains two images and two texts both with minimal
differences in the relation aspect.
The maximum metrics are shown in bold.}
\begin{center}
{\linespread{1.0}
\setlength\tabcolsep{0.5pt}
\scriptsize
\begin{threeparttable}
\begin{tabular}{c | c| c c c  }
\hline
{Dataset} &{Models}
& $Q_{\mathcal{T}:R\&O \xrightarrow{} \mathcal{I}:O_1}$ & $Q_{\mathcal{T}:R\&O \xrightarrow{} \mathcal{I}:O_2}$ & $Q_{\mathcal{T}:R\&O \xrightarrow{} \mathcal{I}:O_1\&O_2}$
\\
\hline
\hline
\multirow{5}{*}{EQBEN}&CLIP   &2.8e-1   & \textbf{6.9e-1}&2.4e-1 \\
&     NEGCLIP &5.1e-1&\textbf{1.3e0} &3.8e-1  \\
&     BLIP &5.1e-1&\textbf{6.7e-1} &4.3e-1 \\
&     XVLM &1.2e-1&\textbf{1.9e-1} &8.5e-2  \\
&     FLAVA &5.4e-1&\textbf{9.2e-1} &6.5e-1\\
\hline
\hline
{Dataset} &{Models}
&  $D_{\mathcal{I}:O_1\&O_2 \xrightarrow{} \mathcal{T}:R}$ & $D_{\mathcal{I}:O_1\&O_2 \xrightarrow{} \mathcal{T}:O}$ & $D_{\mathcal{I}:O_1\&O_2 \xrightarrow{} \mathcal{T}:R\&O}$\\
\hline
\hline
\multirow{5}{*}{EQBEN}&CLIP  & 8.0e-1 &\textbf{2.4e0}   &1.3e0 \\
&     NEGCLIP  &1.0e0&\textbf{3.0e0} &1.3e0  \\
&     BLIP  &6.7e-1&1.5e0 & \textbf{3.6e0}\\
&     XVLM  &1.0e0&1.6e0 &\textbf{3.2e0} \\
&     FLAVA &1.0e0&2.1e0 &\textbf{2.6e0} \\
\hline
\end{tabular}
\end{threeparttable}
}
\end{center}
\label{eqben-relation-t2i}
\end{table}

Similarly, Eq. \ref{eq-di2t} aims to examine VLMs by measuring what components of texts are more related to the interaction between objects on images.
As shown in Figure \ref{fig:4}, we mainly calculated $D_{\mathcal{I}:O_1\&O_2 \xrightarrow{}\mathcal{T}:O/R}$ and $D_{\mathcal{I}:O_1\&O_2 \xrightarrow{}\mathcal{T}:R\&O}$.
Taking $D_{\mathcal{I}:O_1\&O_2 \xrightarrow{}\mathcal{T}:O}$ as an example, it measures how the relation changes between objects within $\mathcal{I}_1$ and $\mathcal{I}_2$ would affect the output of VLMs, when only given the object words in $\mathcal{T}_1$. 
In this case, $D_{\mathcal{I}:O_1\&O_2 \xrightarrow{}\mathcal{T}:R\&O}$ should be of the greatest value, showing that only when the textual pattern demonstrates the relations between objects, altering the relations of objects in images would put significant effects on the output of VLMs.

\noindent \textbf{Experiment results.}
Following previous analyses, we continued analyzing CLIP, NEGCLIP, BLIP, XVLM, FLAVA and harnessed our annotated EQBEN sub-dataset for evaluations, considering each sample contains two images and two texts both with minimal differences.
Results in Tab. \ref{eqben-relation-t2i} show that $Q_{\mathcal{T}:R\&O \xrightarrow{} \mathcal{I}:O_1\&O_2}$ did not maintain the largest among all three metrics, showing that in terms of object relations, image encoders of VLMs did not learn corresponding visual patterns to match the textual object relation patterns encoded inside text encoders. 
Instead, image encoders tended to associate the representations of mere objects with the interactions between object words and relation words learned by text encoders (\emph{e.g.}, $Q_{\mathcal{T}:R\&O \xrightarrow{} \mathcal{I}:O_2}$ showed the largest value for all VLMs.).
Meanwhile, $D_{\mathcal{I}:O_1\&O_2 \xrightarrow{} \mathcal{T}:R\&O}$ failed to maintain the largest among all three metrics across all VLMs as well, showing that regarding the object relations, text encoders also did not associate the corresponding textual patterns to the visual object relation patterns learned by image encoders. 
They sometimes considered the representations of object words alone to be more related to the object relations depicted in images (\emph{e.g.}, $D_{\mathcal{I}:O_1\&O_2 \xrightarrow{} \mathcal{T}:O}$ had the largest value for CLIP/NEGCLIP.).
In summary, \textbf{these models did not exhibit mutually matching compositional knowledge from the text and visual sides, which may also partially account for the poor compositional capabilities of VLMs.}

\section{Conclusion}
\label{sec:conc}
In this paper, we have conducted systematical analyses on the compositionality reasoning capabilities of Vision Language Models (VLMs), which are widely considered as important characteristics of human intelligence.
To this end, we have progressively diagnosed the compositional knowledge of each unimodal encoder \textit{separately} and then the multimodal compositional knowledge \textit{jointly}.
A number of new metrics from a novel game-theoretic view have been proposed to conduct fine-grained compositionality knowledge diagnoses.
In this way, we have obtained and validated several insights regarding the causes for the poor compositional performance of VLMs, which may help provide useful guidance on future explorations.
\section*{Acknowledgements}

This paper is partially supported by the National Key R\&D Program of China No.2022ZD0161000.

\section*{Impact Statement}
This paper presents work whose goal is to diagnose the compositional knowledge of vision language models in a holistic manner.
One positive impact of our work is that our analyses may help instruct future studies to effectively improve the compositional performance of vision language models, so as to advance the development of many downstream unimodal and multimodal applications.
However, it is also crucial to recognize the potential negative impacts, which may result from the malicious misuse of downstream algorithms.

\bibliography{example_paper}

\begin{thebibliography}{74}
\providecommand{\natexlab}[1]{#1}
\providecommand{\url}[1]{\texttt{#1}}
\expandafter\ifx\csname urlstyle\endcsname\relax
  \providecommand{\doi}[1]{doi: #1}\else
  \providecommand{\doi}{doi: \begingroup \urlstyle{rm}\Url}\fi

\bibitem[Ancona et~al.(2019)Ancona, Oztireli, and Gross]{ancona2019explaining}
Ancona, M., Oztireli, C., and Gross, M.
\newblock Explaining deep neural networks with a polynomial time algorithm for shapley value approximation.
\newblock In \emph{International Conference on Machine Learning}, pp.\  272--281. PMLR, 2019.

\bibitem[Cascante-Bonilla et~al.(2023)Cascante-Bonilla, Shehada, Smith, Doveh, Kim, Panda, Varol, Oliva, Ordonez, Feris, et~al.]{cascante2023going}
Cascante-Bonilla, P., Shehada, K., Smith, J.~S., Doveh, S., Kim, D., Panda, R., Varol, G., Oliva, A., Ordonez, V., Feris, R., et~al.
\newblock Going beyond nouns with vision \& language models using synthetic data.
\newblock In \emph{Proceedings of the IEEE/CVF International Conference on Computer Vision}, pp.\  20155--20165, 2023.

\bibitem[Chen et~al.(2023)Chen, Lou, Zhang, Huang, and Zhang]{chen2023harsanyinet}
Chen, L., Lou, S., Zhang, K., Huang, J., and Zhang, Q.
\newblock Harsanyinet: Computing accurate shapley values in a single forward propagation.
\newblock \emph{arXiv preprint arXiv:2304.01811}, 2023.

\bibitem[Chen et~al.(2020)Chen, Li, Yu, El~Kholy, Ahmed, Gan, Cheng, and Liu]{2019UNITER}
Chen, Y.-C., Li, L., Yu, L., El~Kholy, A., Ahmed, F., Gan, Z., Cheng, Y., and Liu, J.
\newblock Uniter: Universal image-text representation learning.
\newblock In \emph{European conference on computer vision}, pp.\  104--120. Springer, 2020.

\bibitem[Cresswell(1973)]{cresswell1973logics}
Cresswell, M.~J.
\newblock Logics and languages.
\newblock 1973.

\bibitem[Deng et~al.(2021)Deng, Ren, Zhang, and Zhang]{deng2021discovering}
Deng, H., Ren, Q., Zhang, H., and Zhang, Q.
\newblock Discovering and explaining the representation bottleneck of dnns.
\newblock \emph{arXiv preprint arXiv:2111.06236}, 2021.

\bibitem[Dong et~al.(2022)Dong, Wang, Liang, Fan, and Ji]{dong2022explaining}
Dong, S., Wang, J., Liang, J., Fan, H., and Ji, R.
\newblock Explaining deepfake detection by analysing image matching.
\newblock In \emph{European Conference on Computer Vision}, pp.\  18--35. Springer, 2022.

\bibitem[Doveh et~al.(2023)Doveh, Arbelle, Harary, Schwartz, Herzig, Giryes, Feris, Panda, Ullman, and Karlinsky]{doveh2023teaching}
Doveh, S., Arbelle, A., Harary, S., Schwartz, E., Herzig, R., Giryes, R., Feris, R., Panda, R., Ullman, S., and Karlinsky, L.
\newblock Teaching structured vision \& language concepts to vision \& language models.
\newblock In \emph{Proceedings of the IEEE/CVF Conference on Computer Vision and Pattern Recognition}, pp.\  2657--2668, 2023.

\bibitem[Du et~al.(2022)Du, Wei, Zhang, Shi, Gao, and Li]{du2022learning}
Du, Y., Wei, F., Zhang, Z., Shi, M., Gao, Y., and Li, G.
\newblock Learning to prompt for open-vocabulary object detection with vision-language model.
\newblock In \emph{Proceedings of the IEEE/CVF Conference on Computer Vision and Pattern Recognition}, pp.\  14084--14093, 2022.

\bibitem[Fan et~al.(2023)Fan, Krishnan, Isola, Katabi, and Tian]{fan2023improving}
Fan, L., Krishnan, D., Isola, P., Katabi, D., and Tian, Y.
\newblock Improving clip training with language rewrites.
\newblock \emph{Advances in Neural Information Processing Systems}, 36, 2023.

\bibitem[Gao et~al.(2022)Gao, Liu, Xu, Zhang, Li, Ji, and Shen]{2022PyramidCLIP}
Gao, Y., Liu, J., Xu, Z., Zhang, J., Li, K., Ji, R., and Shen, C.
\newblock Pyramidclip: Hierarchical feature alignment for vision-language model pretraining.
\newblock \emph{Advances in neural information processing systems}, 35:\penalty0 35959--35970, 2022.

\bibitem[Goel et~al.(2022)Goel, Bansal, Bhatia, Rossi, Vinay, and Grover]{cyclip}
Goel, S., Bansal, H., Bhatia, S., Rossi, R., Vinay, V., and Grover, A.
\newblock Cyclip: Cyclic contrastive language-image pretraining.
\newblock \emph{Advances in Neural Information Processing Systems}, 35:\penalty0 6704--6719, 2022.

\bibitem[Grabisch \& Roubens(1999)Grabisch and Roubens]{grabisch1999axiomatic}
Grabisch, M. and Roubens, M.
\newblock An axiomatic approach to the concept of interaction among players in cooperative games.
\newblock \emph{International Journal of game theory}, 28\penalty0 (4):\penalty0 547--565, 1999.

\bibitem[Grabisch et~al.(2016)]{axioms}
Grabisch, M. et~al.
\newblock \emph{Set functions, games and capacities in decision making}, volume~46.
\newblock Springer, 2016.

\bibitem[Greff et~al.(2022)Greff, Belletti, Beyer, Doersch, Du, Duckworth, Fleet, Gnanapragasam, Golemo, Herrmann, et~al.]{greff2022kubric}
Greff, K., Belletti, F., Beyer, L., Doersch, C., Du, Y., Duckworth, D., Fleet, D.~J., Gnanapragasam, D., Golemo, F., Herrmann, C., et~al.
\newblock Kubric: A scalable dataset generator.
\newblock In \emph{Proceedings of the IEEE/CVF Conference on Computer Vision and Pattern Recognition}, pp.\  3749--3761, 2022.

\bibitem[Gu et~al.(2021)Gu, Lin, Kuo, and Cui]{gu2021open}
Gu, X., Lin, T.-Y., Kuo, W., and Cui, Y.
\newblock Open-vocabulary object detection via vision and language knowledge distillation.
\newblock In \emph{International Conference on Learning Representations}, 2021.

\bibitem[Harsanyi(1963)]{harsanyi}
Harsanyi, J.~C.
\newblock A simplified bargaining model for the n-person cooperative game.
\newblock \emph{International Economic Review}, 4\penalty0 (2):\penalty0 194--220, 1963.

\bibitem[Hertz et~al.(2022)Hertz, Mokady, Tenenbaum, Aberman, Pritch, and Cohen-or]{hertz2022prompt}
Hertz, A., Mokady, R., Tenenbaum, J., Aberman, K., Pritch, Y., and Cohen-or, D.
\newblock Prompt-to-prompt image editing with cross-attention control.
\newblock In \emph{The Eleventh International Conference on Learning Representations}, 2022.

\bibitem[Herzig et~al.(2023)Herzig, Mendelson, Karlinsky, Arbelle, Feris, Darrell, and Globerson]{herzig2023incorporating}
Herzig, R., Mendelson, A., Karlinsky, L., Arbelle, A., Feris, R., Darrell, T., and Globerson, A.
\newblock Incorporating structured representations into pretrained vision \& language models using scene graphs.
\newblock \emph{arXiv preprint arXiv:2305.06343}, 2023.

\bibitem[Honnibal \& Montani(2017)Honnibal and Montani]{spacy}
Honnibal, M. and Montani, I.
\newblock spacy 2: Natural language understanding with bloom embeddings, convolutional neural networks and incremental parsing.
\newblock \emph{To appear}, 7\penalty0 (1):\penalty0 411--420, 2017.

\bibitem[Hsieh et~al.(2023)Hsieh, Zhang, Ma, Kembhavi, and Krishna]{sugarcrepe}
Hsieh, C.-Y., Zhang, J., Ma, Z., Kembhavi, A., and Krishna, R.
\newblock Sugarcrepe: Fixing hackable benchmarks for vision-language compositionality.
\newblock In \emph{Thirty-Seventh Conference on Neural Information Processing Systems Datasets and Benchmarks Track}, 2023.

\bibitem[Huang et~al.(2023{\natexlab{a}})Huang, Sun, Xie, Li, and Liu]{huang2023t2i}
Huang, K., Sun, K., Xie, E., Li, Z., and Liu, X.
\newblock T2i-compbench: A comprehensive benchmark for open-world compositional text-to-image generation.
\newblock In \emph{Thirty-seventh Conference on Neural Information Processing Systems Datasets and Benchmarks Track}, 2023{\natexlab{a}}.

\bibitem[Huang et~al.(2023{\natexlab{b}})Huang, Tang, Chen, Zhang, Zhang, Chen, Zhao, Lv, Hu, and Zhang]{huang2023structure}
Huang, Y., Tang, J., Chen, Z., Zhang, R., Zhang, X., Chen, W., Zhao, Z., Lv, T., Hu, Z., and Zhang, W.
\newblock Structure-clip: Enhance multi-modal language representations with structure knowledge.
\newblock \emph{arXiv preprint arXiv:2305.06152}, 2023{\natexlab{b}}.

\bibitem[Hudson \& Manning(2019)Hudson and Manning]{gqa}
Hudson, D.~A. and Manning, C.~D.
\newblock Gqa: A new dataset for real-world visual reasoning and compositional question answering.
\newblock In \emph{Proceedings of the IEEE/CVF conference on computer vision and pattern recognition}, pp.\  6700--6709, 2019.

\bibitem[Ji et~al.(2020)Ji, Krishna, Fei-Fei, and Niebles]{ji2020action}
Ji, J., Krishna, R., Fei-Fei, L., and Niebles, J.~C.
\newblock Action genome: Actions as compositions of spatio-temporal scene graphs.
\newblock In \emph{Proceedings of the IEEE/CVF Conference on Computer Vision and Pattern Recognition}, pp.\  10236--10247, 2020.

\bibitem[Jia et~al.(2021)Jia, Yang, Xia, Chen, Parekh, Pham, Le, Sung, Li, and Duerig]{align}
Jia, C., Yang, Y., Xia, Y., Chen, Y.-T., Parekh, Z., Pham, H., Le, Q., Sung, Y.-H., Li, Z., and Duerig, T.
\newblock Scaling up visual and vision-language representation learning with noisy text supervision.
\newblock In \emph{International conference on machine learning}, pp.\  4904--4916. PMLR, 2021.

\bibitem[Kirillov et~al.(2023)Kirillov, Mintun, Ravi, Mao, Rolland, Gustafson, Xiao, Whitehead, Berg, Lo, Doll{\'a}r, and Girshick]{SAM}
Kirillov, A., Mintun, E., Ravi, N., Mao, H., Rolland, C., Gustafson, L., Xiao, T., Whitehead, S., Berg, A.~C., Lo, W.-Y., Doll{\'a}r, P., and Girshick, R.
\newblock Segment anything.
\newblock \emph{arXiv:2304.02643}, 2023.

\bibitem[Krishna et~al.(2017)Krishna, Zhu, Groth, Johnson, Hata, Kravitz, Chen, Kalantidis, Li, Shamma, et~al.]{visualgenome}
Krishna, R., Zhu, Y., Groth, O., Johnson, J., Hata, K., Kravitz, J., Chen, S., Kalantidis, Y., Li, L.-J., Shamma, D.~A., et~al.
\newblock Visual genome: Connecting language and vision using crowdsourced dense image annotations.
\newblock \emph{International journal of computer vision}, 123:\penalty0 32--73, 2017.

\bibitem[Li et~al.(2022{\natexlab{a}})Li, Weinberger, Belongie, Koltun, and Ranftl]{li2022languagedriven}
Li, B., Weinberger, K.~Q., Belongie, S., Koltun, V., and Ranftl, R.
\newblock Language-driven semantic segmentation.
\newblock In \emph{International Conference on Learning Representations}, 2022{\natexlab{a}}.
\newblock URL \url{https://openreview.net/forum?id=RriDjddCLN}.

\bibitem[Li et~al.(2021{\natexlab{a}})Li, Selvaraju, Gotmare, Joty, Xiong, and Hoi]{2021Align}
Li, J., Selvaraju, R., Gotmare, A., Joty, S., Xiong, C., and Hoi, S. C.~H.
\newblock Align before fuse: Vision and language representation learning with momentum distillation.
\newblock \emph{Advances in neural information processing systems}, 34:\penalty0 9694--9705, 2021{\natexlab{a}}.

\bibitem[Li et~al.(2022{\natexlab{b}})Li, Li, Xiong, and Hoi]{blip}
Li, J., Li, D., Xiong, C., and Hoi, S.
\newblock Blip: Bootstrapping language-image pre-training for unified vision-language understanding and generation.
\newblock In \emph{International Conference on Machine Learning}, pp.\  12888--12900. PMLR, 2022{\natexlab{b}}.

\bibitem[Li et~al.(2019{\natexlab{a}})Li, Yatskar, Yin, Hsieh, and Chang]{2019VisualBERT}
Li, L.~H., Yatskar, M., Yin, D., Hsieh, C.-J., and Chang, K.-W.
\newblock Visualbert: A simple and performant baseline for vision and language.
\newblock \emph{arXiv preprint arXiv:1908.03557}, 2019{\natexlab{a}}.

\bibitem[Li \& Zhang(2023)Li and Zhang]{li2023does}
Li, M. and Zhang, Q.
\newblock Does a neural network really encode symbolic concept?
\newblock In \emph{{ICML}}, 2023.

\bibitem[Li et~al.(2021{\natexlab{b}})Li, Liang, Zhao, Cui, Ouyang, Shao, Yu, and Yan]{2021Supervision}
Li, Y., Liang, F., Zhao, L., Cui, Y., Ouyang, W., Shao, J., Yu, F., and Yan, J.
\newblock Supervision exists everywhere: A data efficient contrastive language-image pre-training paradigm.
\newblock In \emph{International Conference on Learning Representations}, 2021{\natexlab{b}}.

\bibitem[Li et~al.(2023)Li, Fan, Hu, Feichtenhofer, and He]{li2023scaling}
Li, Y., Fan, H., Hu, R., Feichtenhofer, C., and He, K.
\newblock Scaling language-image pre-training via masking.
\newblock In \emph{Proceedings of the IEEE/CVF Conference on Computer Vision and Pattern Recognition}, pp.\  23390--23400, 2023.

\bibitem[Li et~al.(2019{\natexlab{b}})Li, Xu, Liu, Huang, Xu, Chen, Ma, Wang, Fang, and Lu]{hake}
Li, Y.-L., Xu, L., Liu, X., Huang, X., Xu, Y., Chen, M., Ma, Z., Wang, S., Fang, H.-S., and Lu, C.
\newblock Hake: Human activity knowledge engine.
\newblock \emph{arXiv preprint arXiv:1904.06539}, 2019{\natexlab{b}}.

\bibitem[Liu et~al.(2023)Liu, Li, Wu, and Lee]{liu2023visual}
Liu, H., Li, C., Wu, Q., and Lee, Y.~J.
\newblock Visual instruction tuning, 2023.

\bibitem[Ma et~al.(2023)Ma, Hong, Gul, Gandhi, Gao, and Krishna]{crepe}
Ma, Z., Hong, J., Gul, M.~O., Gandhi, M., Gao, I., and Krishna, R.
\newblock Crepe: Can vision-language foundation models reason compositionally?
\newblock In \emph{Proceedings of the IEEE/CVF Conference on Computer Vision and Pattern Recognition}, pp.\  10910--10921, 2023.

\bibitem[Momeni et~al.(2023)Momeni, Caron, Nagrani, Zisserman, and Schmid]{momeni2023verbs}
Momeni, L., Caron, M., Nagrani, A., Zisserman, A., and Schmid, C.
\newblock Verbs in action: Improving verb understanding in video-language models.
\newblock In \emph{Proceedings of the IEEE/CVF International Conference on Computer Vision}, pp.\  15579--15591, 2023.

\bibitem[OpenAI.(2022)]{chatgpt}
OpenAI.
\newblock Chatgpt.
\newblock 2022.

\bibitem[Peng et~al.(2024)Peng, Xie, You, Lan, and Wu]{peng2024spec}
Peng, W., Xie, S., You, Z., Lan, S., and Wu, Z.
\newblock Synthesize, diagnose, and optimize: Towards fine-grained vision-language understanding.
\newblock In \emph{CVPR}, 2024.

\bibitem[Pham et~al.(2021)Pham, Kafle, Lin, Ding, Cohen, Tran, and Shrivastava]{vaw}
Pham, K., Kafle, K., Lin, Z., Ding, Z., Cohen, S., Tran, Q., and Shrivastava, A.
\newblock Learning to predict visual attributes in the wild.
\newblock In \emph{Proceedings of the IEEE/CVF conference on computer vision and pattern recognition}, pp.\  13018--13028, 2021.

\bibitem[Pratt et~al.(2020)Pratt, Yatskar, Weihs, Farhadi, and Kembhavi]{swig}
Pratt, S., Yatskar, M., Weihs, L., Farhadi, A., and Kembhavi, A.
\newblock Grounded situation recognition.
\newblock In \emph{Computer Vision--ECCV 2020: 16th European Conference, Glasgow, UK, August 23--28, 2020, Proceedings, Part IV 16}, pp.\  314--332. Springer, 2020.

\bibitem[Radford et~al.(2021)Radford, Kim, Hallacy, Ramesh, Goh, Agarwal, Sastry, Askell, Mishkin, Clark, et~al.]{clip}
Radford, A., Kim, J.~W., Hallacy, C., Ramesh, A., Goh, G., Agarwal, S., Sastry, G., Askell, A., Mishkin, P., Clark, J., et~al.
\newblock Learning transferable visual models from natural language supervision.
\newblock In \emph{International conference on machine learning}, pp.\  8748--8763. PMLR, 2021.

\bibitem[Ramesh et~al.(2022)Ramesh, Dhariwal, Nichol, Chu, and Chen]{dalle-2}
Ramesh, A., Dhariwal, P., Nichol, A., Chu, C., and Chen, M.
\newblock Hierarchical text-conditional image generation with clip latents.
\newblock \emph{arXiv preprint arXiv:2204.06125}, 1\penalty0 (2):\penalty0 3, 2022.

\bibitem[Ray et~al.(2023)Ray, Radenovic, Dubey, Plummer, Krishna, and Saenko]{cola}
Ray, A., Radenovic, F., Dubey, A., Plummer, B.~A., Krishna, R., and Saenko, K.
\newblock Cola: How to adapt vision-language models to compose objects localized with attributes?
\newblock \emph{arXiv preprint arXiv:2305.03689}, 2023.

\bibitem[Ren et~al.(2021)Ren, Zhang, Wang, Chen, Zhou, Chen, Cheng, Wang, Zhou, Shi, et~al.]{ren2021unified}
Ren, J., Zhang, D., Wang, Y., Chen, L., Zhou, Z., Chen, Y., Cheng, X., Wang, X., Zhou, M., Shi, J., et~al.
\newblock A unified game-theoretic interpretation of adversarial robustness.
\newblock \emph{arXiv preprint arXiv:2111.03536}, 4, 2021.

\bibitem[Ren et~al.(2022)Ren, Zhou, Chen, and Zhang]{ren2022can}
Ren, J., Zhou, Z., Chen, Q., and Zhang, Q.
\newblock Can we faithfully represent absence states to compute shapley values on a dnn?
\newblock In \emph{The Eleventh International Conference on Learning Representations}, 2022.

\bibitem[Ren et~al.(2023)Ren, Li, Chen, Deng, and Zhang]{ren2023defining}
Ren, J., Li, M., Chen, Q., Deng, H., and Zhang, Q.
\newblock Defining and quantifying the emergence of sparse concepts in dnns.
\newblock In \emph{Proceedings of the IEEE/CVF Conference on Computer Vision and Pattern Recognition}, pp.\  20280--20289, 2023.

\bibitem[Rombach et~al.(2022)Rombach, Blattmann, Lorenz, Esser, and Ommer]{SD}
Rombach, R., Blattmann, A., Lorenz, D., Esser, P., and Ommer, B.
\newblock High-resolution image synthesis with latent diffusion models.
\newblock In \emph{Proceedings of the IEEE/CVF conference on computer vision and pattern recognition}, pp.\  10684--10695, 2022.

\bibitem[Sahin et~al.(2023)Sahin, Li, Khan, Cremers, and Tresp]{sahin2023enhancing}
Sahin, U., Li, H., Khan, Q., Cremers, D., and Tresp, V.
\newblock Enhancing multimodal compositional reasoning of visual language models with generative negative mining.
\newblock \emph{arXiv preprint arXiv:2311.03964}, 2023.

\bibitem[Shapley(1953)]{shapley-value}
Shapley, L.~S.
\newblock A value for n-person games, contributions to the theory of games, 2, 307--317, 1953.

\bibitem[Singh et~al.(2022)Singh, Hu, Goswami, Couairon, Galuba, Rohrbach, and Kiela]{flava}
Singh, A., Hu, R., Goswami, V., Couairon, G., Galuba, W., Rohrbach, M., and Kiela, D.
\newblock Flava: A foundational language and vision alignment model.
\newblock In \emph{Proceedings of the IEEE/CVF Conference on Computer Vision and Pattern Recognition}, pp.\  15638--15650, 2022.

\bibitem[Sundararajan et~al.(2020)Sundararajan, Dhamdhere, and Agarwal]{sundararajan2020shapley}
Sundararajan, M., Dhamdhere, K., and Agarwal, A.
\newblock The shapley taylor interaction index.
\newblock In \emph{International conference on machine learning}, pp.\  9259--9268. PMLR, 2020.

\bibitem[Tan \& Bansal(2019)Tan and Bansal]{2019LXMERT}
Tan, H. and Bansal, M.
\newblock Lxmert: Learning cross-modality encoder representations from transformers.
\newblock In \emph{Proceedings of the 2019 Conference on Empirical Methods in Natural Language Processing and the 9th International Joint Conference on Natural Language Processing (EMNLP-IJCNLP)}, pp.\  5100--5111, 2019.

\bibitem[Thrush et~al.(2022)Thrush, Jiang, Bartolo, Singh, Williams, Kiela, and Ross]{winoground}
Thrush, T., Jiang, R., Bartolo, M., Singh, A., Williams, A., Kiela, D., and Ross, C.
\newblock Winoground: Probing vision and language models for visio-linguistic compositionality.
\newblock In \emph{Proceedings of the IEEE/CVF Conference on Computer Vision and Pattern Recognition}, pp.\  5238--5248, 2022.

\bibitem[Wang et~al.(2023)Wang, Lin, Li, Lin, Yang, Zhang, Liu, and Wang]{eqben}
Wang, T., Lin, K., Li, L., Lin, C.-C., Yang, Z., Zhang, H., Liu, Z., and Wang, L.
\newblock Equivariant similarity for vision-language foundation models.
\newblock In \emph{Proceedings of the IEEE/CVF International Conference on Computer Vision (ICCV)}, pp.\  11998--12008, October 2023.

\bibitem[Wang et~al.(2020)Wang, Ren, Lin, Zhu, Wang, and Zhang]{wang2020unified}
Wang, X., Ren, J., Lin, S., Zhu, X., Wang, Y., and Zhang, Q.
\newblock A unified approach to interpreting and boosting adversarial transferability.
\newblock In \emph{International Conference on Learning Representations}, 2020.

\bibitem[Wang et~al.(2021)Wang, Lin, Zhang, Zhu, and Zhang]{wang2021interpreting}
Wang, X., Lin, S., Zhang, H., Zhu, Y., and Zhang, Q.
\newblock Interpreting attributions and interactions of adversarial attacks.
\newblock In \emph{Proceedings of the IEEE/CVF International Conference on Computer Vision}, pp.\  1095--1104, 2021.

\bibitem[Wang et~al.(2022)Wang, Gao, Yu, Lei, Feiszli, and Shou]{wang2022geb+}
Wang, Y., Gao, D., Yu, L., Lei, W., Feiszli, M., and Shou, M.~Z.
\newblock Geb+: A benchmark for generic event boundary captioning, grounding and retrieval.
\newblock In \emph{European Conference on Computer Vision}, pp.\  709--725. Springer, 2022.

\bibitem[Xu et~al.(2022)Xu, De~Mello, Liu, Byeon, Breuel, Kautz, and Wang]{xu2022groupvit}
Xu, J., De~Mello, S., Liu, S., Byeon, W., Breuel, T., Kautz, J., and Wang, X.
\newblock Groupvit: Semantic segmentation emerges from text supervision.
\newblock In \emph{Proceedings of the IEEE/CVF Conference on Computer Vision and Pattern Recognition}, pp.\  18134--18144, 2022.

\bibitem[Yao et~al.(2023)Yao, Wang, Diao, and Li]{yao2023towards}
Yao, K., Wang, J., Diao, B., and Li, C.
\newblock Towards understanding the generalization of deepfake detectors from a game-theoretical view.
\newblock In \emph{Proceedings of the IEEE/CVF International Conference on Computer Vision}, pp.\  2031--2041, 2023.

\bibitem[Yao et~al.(2021)Yao, Huang, Hou, Lu, Niu, Xu, Liang, Li, Jiang, and Xu]{filip}
Yao, L., Huang, R., Hou, L., Lu, G., Niu, M., Xu, H., Liang, X., Li, Z., Jiang, X., and Xu, C.
\newblock Filip: Fine-grained interactive language-image pre-training.
\newblock \emph{arXiv preprint arXiv:2111.07783}, 2021.

\bibitem[Yuksekgonul et~al.(2022)Yuksekgonul, Bianchi, Kalluri, Jurafsky, and Zou]{aro}
Yuksekgonul, M., Bianchi, F., Kalluri, P., Jurafsky, D., and Zou, J.
\newblock When and why vision-language models behave like bags-of-words, and what to do about it?
\newblock In \emph{The Eleventh International Conference on Learning Representations}, 2022.

\bibitem[Zang et~al.(2022)Zang, Li, Zhou, Huang, and Loy]{zang2022open}
Zang, Y., Li, W., Zhou, K., Huang, C., and Loy, C.~C.
\newblock Open-vocabulary detr with conditional matching.
\newblock In \emph{European Conference on Computer Vision}, pp.\  106--122. Springer, 2022.

\bibitem[Zeng et~al.(2022)Zeng, Zhang, and Li]{xvlm}
Zeng, Y., Zhang, X., and Li, H.
\newblock Multi-grained vision language pre-training: Aligning texts with visual concepts.
\newblock In \emph{International Conference on Machine Learning}, pp.\  25994--26009. PMLR, 2022.

\bibitem[Zhang et~al.(2021{\natexlab{a}})Zhang, Zhang, Zhou, Bao, Huo, Chen, Cheng, Wu, and Zhang]{zhang2021building}
Zhang, D., Zhang, H., Zhou, H., Bao, X., Huo, D., Chen, R., Cheng, X., Wu, M., and Zhang, Q.
\newblock Building interpretable interaction trees for deep nlp models.
\newblock In \emph{Proceedings of the AAAI Conference on Artificial Intelligence}, volume~35, pp.\  14328--14337, 2021{\natexlab{a}}.

\bibitem[Zhang et~al.(2020{\natexlab{a}})Zhang, Cheng, Chen, and Zhang]{zhang2020game}
Zhang, H., Cheng, X., Chen, Y., and Zhang, Q.
\newblock Game-theoretic interactions of different orders.
\newblock \emph{arXiv preprint arXiv:2010.14978}, 2020{\natexlab{a}}.

\bibitem[Zhang et~al.(2020{\natexlab{b}})Zhang, Li, Ma, Li, Xie, and Zhang]{zhang2020interpreting}
Zhang, H., Li, S., Ma, Y., Li, M., Xie, Y., and Zhang, Q.
\newblock Interpreting and boosting dropout from a game-theoretic view.
\newblock In \emph{International Conference on Learning Representations}, 2020{\natexlab{b}}.

\bibitem[Zhang et~al.(2021{\natexlab{b}})Zhang, Xie, Zheng, Zhang, and Zhang]{zhang2021interpreting}
Zhang, H., Xie, Y., Zheng, L., Zhang, D., and Zhang, Q.
\newblock Interpreting multivariate shapley interactions in dnns.
\newblock In \emph{Proceedings of the AAAI Conference on Artificial Intelligence}, volume~35, pp.\  10877--10886, 2021{\natexlab{b}}.

\bibitem[Zhao et~al.(2022)Zhao, Zhang, Zhu, Shen, Lee, Lu, and Yin]{vlchecklist}
Zhao, T., Zhang, T., Zhu, M., Shen, H., Lee, K., Lu, X., and Yin, J.
\newblock Vl-checklist: Evaluating pre-trained vision-language models with objects, attributes and relations, 2022.
\newblock URL \url{https://arxiv.org/abs/2207.00221}.

\bibitem[Zhou et~al.(2022)Zhou, Loy, and Dai]{zhou2022extract}
Zhou, C., Loy, C.~C., and Dai, B.
\newblock Extract free dense labels from clip.
\newblock In \emph{European Conference on Computer Vision}, pp.\  696--712. Springer, 2022.

\bibitem[Zhou et~al.(2018)Zhou, Xu, and Corso]{zhou2018towards}
Zhou, L., Xu, C., and Corso, J.
\newblock Towards automatic learning of procedures from web instructional videos.
\newblock In \emph{Proceedings of the AAAI Conference on Artificial Intelligence}, volume~32, 2018.

\bibitem[Zhu et~al.(2023)Zhu, Chen, Shen, Li, and Elhoseiny]{zhu2023minigpt4}
Zhu, D., Chen, J., Shen, X., Li, X., and Elhoseiny, M.
\newblock Minigpt-4: Enhancing vision-language understanding with advanced large language models, 2023.

\end{thebibliography}
\bibliographystyle{icml2024}

%%%%%%%%%%%%%%%%%%%%%%%%%%%%%%%%%%%%%%%%%%%%%%%%%%%%%%%%%%%%%%%%%%%%%%%%%%%%%%%
%%%%%%%%%%%%%%%%%%%%%%%%%%%%%%%%%%%%%%%%%%%%%%%%%%%%%%%%%%%%%%%%%%%%%%%%%%%%%%%
% APPENDIX
%%%%%%%%%%%%%%%%%%%%%%%%%%%%%%%%%%%%%%%%%%%%%%%%%%%%%%%%%%%%%%%%%%%%%%%%%%%%%%%
%%%%%%%%%%%%%%%%%%%%%%%%%%%%%%%%%%%%%%%%%%%%%%%%%%%%%%%%%%%%%%%%%%%%%%%%%%%%%%%
\newpage
\appendix
\onecolumn
\section{Axioms of the Harsanyi dividend}
\label{sec:rationale}
The Harsanyi dividend \cite{harsanyi} satisfies many axioms, which provide solid theoretical foundations for our explanations in the main paper.
The axioms are as follows.
\begin{itemize}
\item[$\bullet$] \textit{Linearity axiom.} Given a game $t$ combined by a game $u$ and a game $v$, \emph{i.e.}, $t(\cdot) = u(\cdot) + v(\cdot)$, the Harsanyi dividend of any subset of players $\mathcal{S}$ in the game $t$ is equal to the sum of the Harsanyi dividends in the game $u$ and $v$, \emph{i.e.}, $w_t({\mathcal{S}}|\mathcal{N}) = w_u({\mathcal{S}}|\mathcal{N}) + w_v({\mathcal{S}}|\mathcal{N})$.
\item[$\bullet$] \textit{Dummy axiom.} If $\forall \mathcal{S}\subseteq \mathcal{N} \backslash \{i\}, v(\mathcal{S}\cup \{i\}) = v(\mathcal{S}) + v(\{i\})$, then player $i$ is a dummy player, having no interactions with other players,\emph{i.e.}, $w({\mathcal{S} \cup \{i \}}|\mathcal{N}) = 0$.
\item[$\bullet$] \textit{Symmetry axiom.} Given two players $i$ and $j$, if $\forall \mathcal{S}\subseteq \mathcal{N} \backslash \{i,j\}, v(\mathcal{S}\cup \{i\}) = v(\mathcal{N}\cup \{j\})$, then they have the same interaction effects with other players, \emph{i.e.}, $\forall \mathcal{S}\subseteq \mathcal{N} \backslash \{i,j\}, w(\mathcal{S}\cup \{i\}|\mathcal{N}) = w(\mathcal{S}\cup \{j\}|\mathcal{N})$.
\item[$\bullet$] \textit{Efficiency axiom.} The overall output of the game $v(\mathcal{N})$ can be disentangled into the interaction effects of different subsets of players $\mathcal{S}$, \emph{i.e.}, $v(\mathcal{N})  = \sum_{\mathcal{S}\subseteq \mathcal{N}} w(\mathcal{S}| \mathcal{N})$.
\end{itemize}

Besides, the Harsanyi dividend is also related to the Shapley value \cite{shapley-value}, as described in the following theorem.

\noindent \textbf{Theorem 1} ({proven in} \cite{harsanyi,ren2022can}). \textit{Let $\phi(i)$ represents the Shapley value of the player $i \in \mathcal{N}$. 
In this way, we have $
\phi(i)=\sum_{\mathcal{S} \subseteq \mathcal{N} \backslash\{i\}} \frac{1}{|S|+1} w({S \cup\{i\}}|\mathcal{N})$, showing that the Shapley value can be considered as the uniform allocations from the numerical values of Harsanyi dividends.
}

\section{Visualizing the new annotations of EQBEN}
\label{supp:eqben:vis}
\begin{figure*}[t]
\centering
\includegraphics [width=0.99\textwidth]{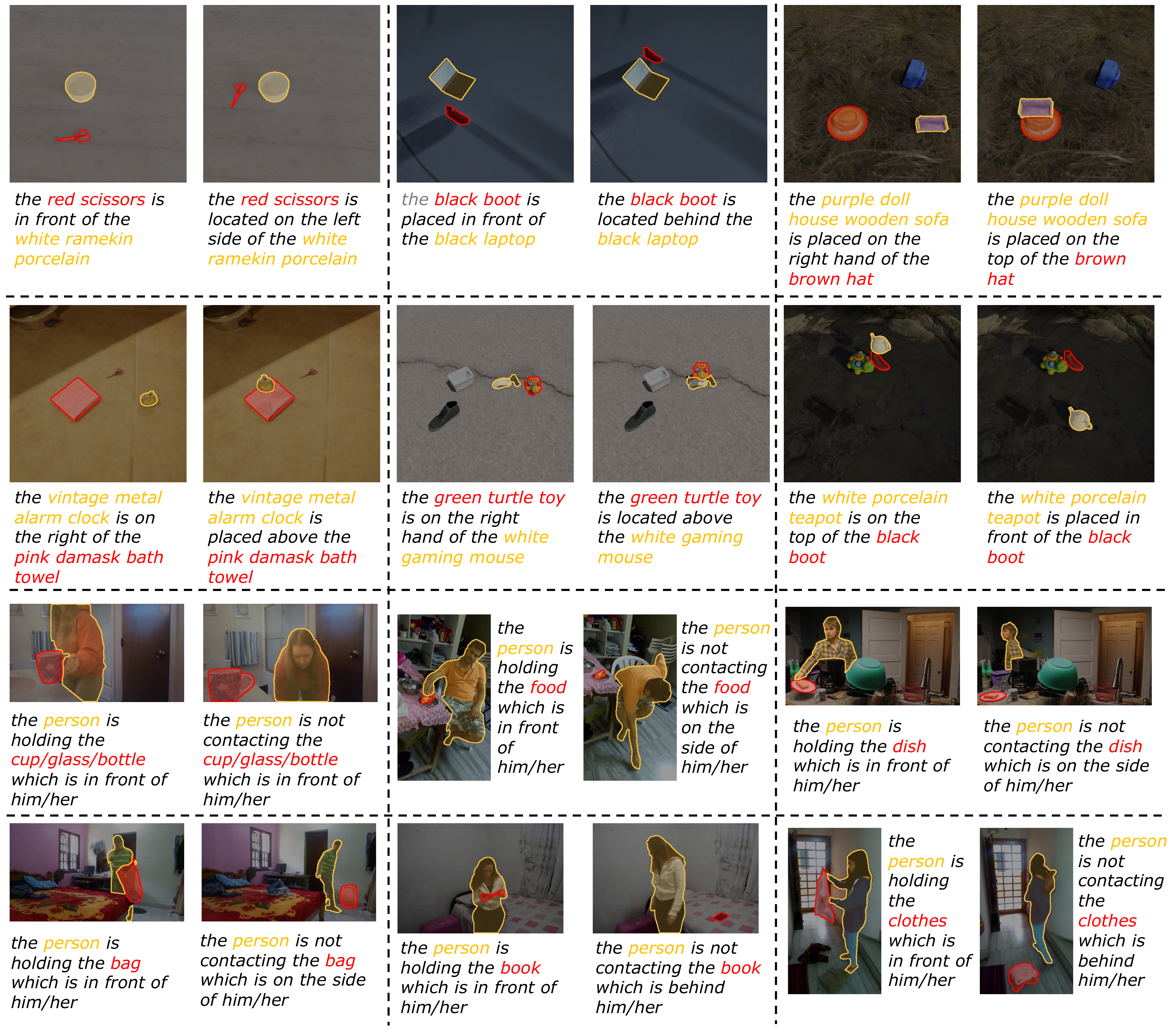}
\caption{
\textbf{Visualizing the new annotations on the subset of the EQBEN dataset.}
In this paper, we leveraged SAM \cite{SAM} to help obtain segmentation results for objects on images. Specifically, we first manually annotated the bounding box for each object described in the corresponding caption. 
We then harnessed SAM to obtain the segmentation mask for each object.
Unsatisfying segmentation results were manually annotated afterward for corrections.
}
\label{fig:bb}
\end{figure*}

In this section, we provide samples from our new object annotations in the subset of the EQBEN dataset. 
The visualization results are presented in Figure \ref{fig:bb}. 
%--------- figure 3. --------------

\section{More results on diagnosing the compositional knowledge of text encoders}
\label{supp:text_side}

\textbf{Experimental results on Visual Genome Attribution dataset.} In this dataset, the correct captions and wrong captions are in the templates of \textit{``the
[attribute 1] [object 1] and the [attribute 2] [object 2]”} and \textit{``the [attribute 2] [object 1] and
the [attribute 1] [object 2]"} with the \textit{SWAP} manipulation. 
Therefore, we mainly measured the sensitivity metrics of attribute words, object words and the interactions between attribute words and object words, \emph{i.e.}, $Q_A$, $Q_O$ and $Q_{A\&O}$.
Masks denoting the attribute words and object words were obtained based on the templates.
Intuitively, this type of caption pair is majorly different in the aspect of the interaction between attribute words and object words, rather than attribute/object words alone. 
Results in Table \ref{tab:aro-attribute} are consistent with this human intuition, showing $Q_{A\&O}$ has the largest value among other metrics.
\textbf{Such results indicate that text encoders also recognize the prominent compositional differences between captions in the attribute-object aspect.}

\begin{figure*}[t]
\centering
\includegraphics[width=1.0\textwidth]{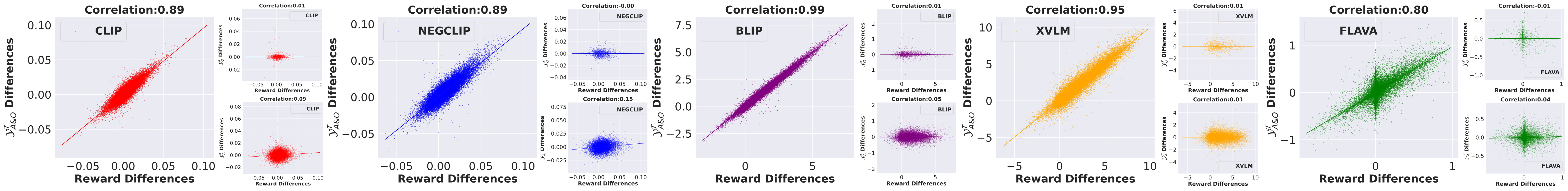}
\caption{The Pearson correlation coefficients $\rho(\mathcal{X^T}, \mathcal{Y^T})$ between the reward differences $\mathcal{X^T}$ and interaction effect differences on the Visual Genome Attribution dataset, \emph{i.e.}, $\mathcal{Y}^\mathcal{T}_A$, $\mathcal{Y}^\mathcal{T}_O$ and $\mathcal{Y}^\mathcal{T}_{A \& O}$.
Each point represents a data sample containing two captions and one image.
Results show that the reward differences between the two captions were mainly related to the interaction changes of object words and attribute words.}
\label{fig:rho:aro-attrbute}
\end{figure*}

Besides, we also calculated the Pearson correlation coefficients $\rho (\mathcal{X^T}, \mathcal{Y^T})$ between reward differences and interaction effect differences in the Visual Genome Attribution dataset, \emph{i.e.}, $\rho (\mathcal{X^T}, \mathcal{Y}^\mathcal{T}_{O})$, $\rho (\mathcal{X^T}, \mathcal{Y}^\mathcal{T}_{A})$, $\rho (\mathcal{X^T}, \mathcal{Y}^\mathcal{T}_{A\&O})$.
$\mathcal{Y}^\mathcal{T}_{A}$ represents the casual effect changes when $\mathcal{S}^{\mathcal{T}_{1}}$ and  $\mathcal{S}^{\mathcal{T}_{2}}$ only contain attribute words.
$\mathcal{Y}^\mathcal{T}_{A\&O}$ represents the casual effect changes when $\mathcal{S}^{\mathcal{T}_{1}}$ and  $\mathcal{S}^{\mathcal{T}_{2}}$ contain both object and attribute words.
Results in Figure \ref{fig:rho:aro-attrbute} show that the interaction effect changes between attribute words and object words played a major part in the final reward differences, showing that text encoders correctly reflected the prominent compositional differences within each caption pair.

\begin{table}[t]
\caption{Evaluating the compositional sensitivities of text encoders of VLMs.
In the Visual Genome Attribution dataset, the attribute words are swapped to obtain perturbed texts. 
Results show that $Q_{A\&O}$ (bold) have larger values than $Q_{A}$ and $Q_{O}$ in this dataset across different VLMs, showing that text encoders of various VLMs exhibit accurate sensibilities to the changes of textual patterns.}
\label{tab:aro-attribute}
\begin{center}
{\linespread{1.0}
\setlength\tabcolsep{6pt}
\scriptsize
\begin{threeparttable}
\begin{tabular}{l l c c c}
\hline
{Dataset} &{Models}
& $Q_O$ & $Q_A$ & $Q_{A \& O}$ \\
\hline
\hline
\multirow{5}{*}{Visual Genome Attribution}&CLIP& 1.2e-4 & 2.9e-3 & \textbf{1.6e-2}  \\
&NEGCLIP& 2.6e-4&4.5e-3 &\textbf{2.9e-2} \\
&BLIP&3.0e-3&2.5e-2 &\textbf{1.9e0}\\
&XVLM&3.3e-3&2.0e-2 &\textbf{4.5e-1}\\
&FLAVA&2.0e-2		
&9.6e-2 & \textbf{2.9e-1} \\
\hline
\end{tabular}
\end{threeparttable}
}
\end{center}

\end{table}

\noindent \textbf{Experimental results on VL-CheckList benchmark.} 
The VL-CheckList benchmark \cite{vlchecklist} contains a large scale of images and captions (over 100K) combined from 4 datasets: Visual Genome \cite{visualgenome}, SWiG \cite{swig}, VAW \cite{vaw}, and HAKE \cite{hake} datasets.
In each sample, there exist one image, one correct caption and one wrong caption.
The wrong caption is generated by \textit{REPLACING} one compositionality aspect of the correct caption, including objects, relations and attributes. 
Furthermore, these samples were then divided into 9 categories: 1) Object: the location and size of it, 2) Relation: action or spatial relation between objects, 3) Attribute: color, material, size, state and action.
In experiments, we exploited Spacy \cite{spacy} for part-of-speech tagging to roughly divide each sentence into object words, relation words and attribute words.
Other irrelevant words were then treated as the constant text background.

We report the results in Table \ref{vlchecklist-object} and Table \ref{vlchecklist-attribute}. 
Intuitively, by replacing the correct caption with new object/relation/attribute words, not only does the effect of object/relation/attribute words change, but also the interactions among them should vary significantly.
As shown in Table \ref{vlchecklist-object}, on the object aspect,  $Q_{R\&O}$ and $Q_O$ (as well as $\rho_{R\&O}$ and $\rho_O$) had the highest values across different text encoders of VLMs, demonstrating their capabilities to recognize the replacing effect of object words. 
Besides, $Q_O$ usually had a slightly higher value than $Q_{R\&O}$, showing that text encoders considered that the replacement of object words affected less on the relation changes between objects.

On the relation aspect in Table \ref{vlchecklist-object}, results showed that $Q_R$ and $Q_{R\&O}$ (as well as $\rho_{R}$ and $\rho_{R\&O}$) had the largest values, showing that text encoders of VLMs recognized the changes of relation words and the interaction changes between relation words and object words.
Besides, for action relations, results show that text encoders of CLIP and NEGCLIP considered that the replacement of relation words caused less impact on the interaction changes between object words and relation words (\emph{i.e.}, $Q_{R\&O}$/$\rho_{R\&O}$ having a smaller value than $Q_{R}$/$\rho_{R}$).
Meanwhile, text encoders of BLIP, XVLM and FLAVA had opposite understandings on the replacement of relation words (\emph{i.e.}, $Q_{R\&O}$/$\rho_{R\&O}$ having a larger value than $Q_{R}$/$\rho_{R}$).
As for spatial relations, all text encoders of these VLMs considered the replacement of relation words affected more on interaction changes between object words and relation words.

In Table \ref{vlchecklist-attribute}, we present the results on the attribute aspect in the VL-CheckList dataset.
Similar to the results in Table \ref{vlchecklist-object}, $Q_A$ and $Q_{A\&O}$ (as well as $\rho_{A}$ and $\rho_{A\&O}$) had the largest values in general, indicating that text encoders of VLMs successfully recognized the changes of attribute words and the interaction changes between attribute words and object words.
Besides, it is noteworthy that for replacing attribute words in captions, text encoders of CLIP and NEGCLIP usually paid more attention to the changes of attribute words alone (\emph{i.e.}, $Q_{A\&O}$/$\rho_{A\&O}$ having a smaller value than $Q_{A}$/$\rho_{A}$), which may expose certain deficiency of CLIP on understanding the close binding relationship between attribute words and object words.
Such results may provide certain explanations for the phenomenon that Stable Diffusion models \cite{SD} struggled to generate correct images given the type of attribute binding text prompts \cite{huang2023t2i}, since they learned their text encoders from CLIP.

In summary, results in Table \ref{vlchecklist-object} and Table \ref{vlchecklist-attribute} indicate that text encoders of VLMs recognized the effects of replacing words in texts similar to human understanding.
Besides, experimental results on such a large-scale dataset further strengthen the trustworthiness of our explanations on diagnosing the compositional reasoning capabilities of text encoders.

\begin{table}[t]
\caption{Evaluating the compositional knowledge encoded in text encoders of VLMs with the VL-CheckList dataset (object/relation aspect). 
Here $\rho_{O}$, $\rho_{R}$ and  $\rho_{R\&O}$ represents $\rho (\mathcal{X^T}, \mathcal{Y}^\mathcal{T}_{O})$, $\rho (\mathcal{X^T}, \mathcal{Y}^\mathcal{T}_{R})$ and $\rho (\mathcal{X^T}, \mathcal{Y}^\mathcal{T}_{R\&O})$ for short.
The top two largest metrics are in bold.
On the object aspect, results show that $Q_O$ and $Q_{R\& O}$ (as well as $\rho_{O}$ and $\rho_{R\&O}$) had the largest values, showing that text encoders of VLMs recognized both the changes of object words and the interaction changes between object words and relation words, when replacing the object words in captions.
On the relation aspect, results show that $Q_R$ and $Q_{R\& O}$ (as well as $\rho_{R}$ and $\rho_{R\&O}$) had the largest values, showing that when replacing the relation words in captions, text encoders of VLMs successfully recognized both the changes of relation words and the interaction changes between relation words and object words.
}
\begin{center}
{\linespread{1.0}
\setlength\tabcolsep{2pt}
\scriptsize
\begin{threeparttable}
\begin{tabular}{c|c | c| c c c |c c c }
\hline
\makecell[c]{Dataset} &Category &{Models}
& $Q_O$ & $Q_R$ & $Q_{R \& O}$
& $\rho_O$ & $\rho_R$ & $\rho_{R\&O}$\\
\hline
\hline
\multirow{10}{*}{\makecell[c]{VL-CheckList \\ (Object)}}& \multirow{5}{*}{Location} &CLIP  &\textbf{3.7e-1}   & 1.1e-2  &\textbf{3.1e-2}&\textbf{0.86} & -0.01&\textbf{0.22} \\
&    & NEGCLIP &\textbf{3.6e-1}&6.5e-3 & \textbf{2.1e-2}&\textbf{0.87}&0.00& \textbf{0.18} \\
&    & BLIP &\textbf{3.8e0}&4.3e-2 &\textbf{9.3e-1} &\textbf{0.65}&-0.05&\textbf{0.44} \\
&    & XVLM &\textbf{1.4e0}&9.7e-2 & \textbf{2.9e-1}&\textbf{0.55}&0.03&\textbf{0.28} \\
&    & FLAVA &\textbf{2.5e0}&3.3e-1 & \textbf{1.0e0}&\textbf{0.43}&0.05&\textbf{0.29} \\
\cline{2-9}
& \multirow{5}{*}{Size} &CLIP  & \textbf{3.2e-1} & 1.5e-2  &\textbf{3.6e-2}&\textbf{0.81} & 0.00& \textbf{0.22}\\
&   & NEGCLIP &\textbf{3.2e-1}&6.0e-3 &\textbf{2.0e-2} &\textbf{0.88}&0.01&\textbf{0.18}  \\
&    & BLIP &\textbf{3.7e0}&4.4e-2 &\textbf{9.0e-1} &\textbf{0.66}&-0.04&\textbf{0.44} \\
&    & XVLM &\textbf{1.5e0}& 9.6e-2& \textbf{2.9e-1}&\textbf{0.55}&0.03&\textbf{0.27} \\
&    & FLAVA &\textbf{2.3e0}&3.4e-1 & \textbf{1.0e0}&\textbf{0.44}&0.05&\textbf{0.28} \\
\hline
\multirow{10}{*}{\makecell[c]{VL-CheckList \\ (Relation)}}& \multirow{5}{*}{Action} &CLIP  & 4.5e-2 & \textbf{1.0e-1}  &\textbf{5.0e-2}&0.02 & \textbf{0.46}&\textbf{0.21}  \\
&    & NEGCLIP &1.8e-2&\textbf{9.2e-2} &\textbf{3.5e-2} &-0.01&\textbf{0.56}&\textbf{0.32}  \\
&    & BLIP &3.0e-1& \textbf{7.3e-1}&\textbf{1.5e0} &-0.02&\textbf{0.36}&\textbf{0.61} \\
&    & XVLM &2.6e-1& \textbf{3.0e-1}&\textbf{4.1e-1} &-0.05&\textbf{0.17}&\textbf{0.47} \\
&    & FLAVA &3.9e-1& \textbf{8.6e-1}&\textbf{1.5e0} &0.16&\textbf{0.19}&\textbf{0.51} \\
\cline{2-9}
& \multirow{5}{*}{Spatial} &CLIP  &1.5e-2  &\textbf{4.3e-2}  &\textbf{5.4e-2}&0.02 &\textbf{0.11}&\textbf{0.53}  \\
&   & NEGCLIP &9.2e-3&\textbf{1.3e-2} &\textbf{2.1e-2} &0.03&\textbf{0.23}&\textbf{0.64}  \\
&    & BLIP &1.7e-1&\textbf{2.3e-1} &\textbf{9.7e-1} &0.01&\textbf{0.12}&\textbf{0.79} \\
&    & XVLM &9.3e-2&\textbf{2.2e-1} &\textbf{2.9e-1} &-0.01&\textbf{0.15}&\textbf{0.43} \\
&    & FLAVA &2.6e-1&\textbf{6.7e-1} &\textbf{7.2e-1} &-0.05&\textbf{0.29}&\textbf{0.51} \\
\hline
\end{tabular}
\end{threeparttable}
}
\end{center}
\label{vlchecklist-object}
\end{table}

\begin{table}[t]
\caption{Evaluating the compositional knowledge encoded in text encoders of VLMs with the VL-CheckList dataset (attribute aspect).
Here $\rho_{O}$, $\rho_{A}$ and  $\rho_{A\&O}$ represents $\rho (\mathcal{X^T}, \mathcal{Y}^\mathcal{T}_{O})$, $\rho (\mathcal{X^T}, \mathcal{Y}^\mathcal{T}_{A})$ and $\rho (\mathcal{X^T}, \mathcal{Y}^\mathcal{T}_{A\&O})$ for short.
The top two largest metrics are in
bold.
Results show that $Q_A$ and $Q_{A\&O}$ (as well as $\rho_{A}$ and  $\rho_{A\&O}$) had the largest values in general, indicating that when replacing the attribute words in captions, text encoders successfully recognized both the changes of attribute words and the interaction changes between object words and attribute words.
}
\begin{center}
{\linespread{1.0}
\setlength\tabcolsep{2pt}
\scriptsize
\begin{threeparttable}
\begin{tabular}{c|c | c| c c c |c c c }
\hline
\makecell[c]{Dataset} &Category &{Models}
& $Q_O$ & $Q_A$ & $Q_{A \& O}$
& $\rho_O$ & $\rho_A$ & $\rho_{A\&O}$\\
\hline
\hline
\multirow{25}{*}{\makecell[c]{VL-CheckList \\ (Attribute)}}& \multirow{5}{*}{Color} &CLIP  &  4.4e-3  & \textbf{ 7.6e-2 }  &\textbf{4.0e-2 }&0.02 &\textbf{0.65}& \textbf{0.30} \\
&    & NEGCLIP &3.1e-3&\textbf{8.5e-2} & \textbf{2.0e-2}&0.00&\textbf{0.81}& \textbf{0.05}\\
&    & BLIP &4.9e-2&\textbf{7.7e-1} &\textbf{2.3e0} &0.04&\textbf{0.45}&\textbf{0.84} \\
&    & XVLM &7.8e-2 &\textbf{3.4e-1}  & \textbf{6.2e-1}&0.02&\textbf{0.18}& \textbf{0.59}\\
&    & FLAVA &1.4e-1& \textbf{8.1e-1}&\textbf{1.1e0} &0.14&\textbf{0.16}&\textbf{0.46} \\
\cline{2-9}
& \multirow{5}{*}{Material} &CLIP  & 1.3e-2 &   \textbf{8.6e-2 }&\textbf{4.8e-2}& -0.03&\textbf{0.58}&\textbf{0.35}  \\
&   & NEGCLIP &3.6e-3&\textbf{9.2e-2} & \textbf{3.0e-2}&0.00&\textbf{0.70}& \textbf{0.13} \\
&    & BLIP &1.0e-1&\textbf{1.3e0} &\textbf{1.9e0} &0.11&\textbf{0.42}&\textbf{0.71} \\
&    & XVLM &1.5e-1 & \textbf{5.6e-1}& \textbf{4.5e-1} &0.05&\textbf{0.17}& \textbf{0.39}\\
&    & FLAVA &3.0e-1&\textbf{1.0e0} &\textbf{1.2e0} &0.21&\textbf{0.35}&\textbf{0.41} \\
\cline{2-9}
& \multirow{5}{*}{Size} &CLIP &1.1e-2 &\textbf{3.6e-2}  &  \textbf{3.6e-2} &0.06&\textbf{0.32} & \textbf{0.49} \\
&   & NEGCLIP &6.1e-3 &\textbf{4.0e-2} & \textbf{2.5e-2}&0.03&\textbf{0.53}&\textbf{0.23}  \\
&    & BLIP &4.8e-2&\textbf{6.5e-1} &\textbf{1.2e0} &0.03&\textbf{0.33}&\textbf{0.77} \\
&    & XVLM &4.5e-2 &\textbf{2.9e-1}  &\textbf{3.0e-1}  &0.04&\textbf{0.23}& \textbf{0.51}\\
&    & FLAVA &9.0e-2&\textbf{4.3e-1} &\textbf{8.7e-1} &-0.04&\textbf{0.19}&\textbf{0.46} \\
\cline{2-9}
& \multirow{5}{*}{State} &CLIP  & 9.1e-3 & \textbf{4.4e-2}  &\textbf{3.5e-2}&0.04 & \textbf{0.50}&\textbf{0.48} \\
&   & NEGCLIP &4.7e-3 & \textbf{6.8e-2} & \textbf{2.3e-2}&0.04&\textbf{0.71}& \textbf{0.16} \\
&    & BLIP &4.0e-2&\textbf{9.9e-1} &\textbf{1.0e0} &-0.03&\textbf{0.55}&\textbf{0.73} \\
&    & XVLM &8.1e-2 &\textbf{4.6e-1}  &\textbf{3.6e-1} &-0.01&\textbf{0.22}& \textbf{0.44}\\
&    & FLAVA &1.7e-1&\textbf{8.1e-1} &\textbf{1.2e0} &0.06&\textbf{0.18}&\textbf{0.42} \\
\cline{2-9}
& \multirow{5}{*}{Action} &CLIP  &1.8e-2   & \textbf{1.4e-1 }  &\textbf{4.8e-2}& 0.01&\textbf{0.68}& \textbf{0.10}  \\
&   & NEGCLIP &1.4e-2 & \textbf{1.7e-1} &\textbf{5.5e-2} &\textbf{0.07}&\textbf{0.72}& -0.05 \\
&    & BLIP &1.1e-1 &\textbf{1.7e0} &\textbf{9.6e-1}&-0.02&\textbf{0.64}&\textbf{0.57} \\
&    & XVLM &1.2e-1 &\textbf{7.2e-1}  &\textbf{4.0e-1} &-0.01&\textbf{0.27}& \textbf{0.33}\\
&    & FLAVA &2.1e-1&\textbf{1.1e0} &\textbf{1.2e0} &-0.01&\textbf{0.15}&\textbf{0.38} \\
\hline
\end{tabular}
\end{threeparttable}
}
\end{center}
\label{vlchecklist-attribute}
\end{table}

%--------- figure 3. --------------
\begin{figure*}[t]
\centering
\includegraphics [width=0.99\textwidth]{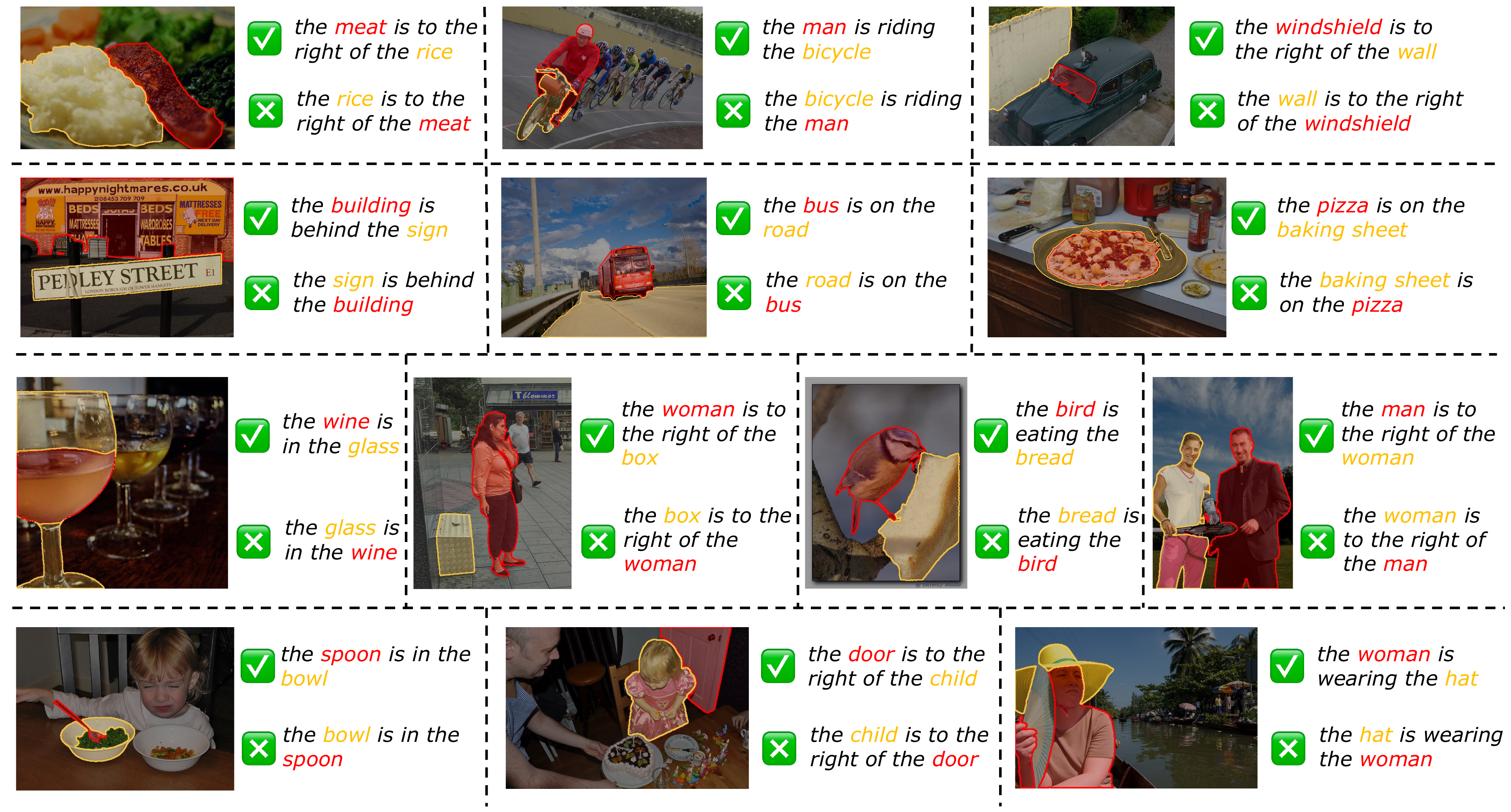}
\caption{
\textbf{Visualizing the new annotations on the subset of the Visual Genome Relation dataset.}
Specifically, we first exploited the original bounding box annotations from the GQA dataset \cite{gqa} and then used SAM \cite{SAM} to obtain the segmentation mask for each object.
We then manually annotated unsatisfying segmentation results afterward for corrections.
}
\label{fig:cc}
\end{figure*}

\noindent \textbf{Experimental results on SUGARCREPE benchmark.} 
The SUGARCREPE benchmark \cite{sugarcrepe} contains a significant amount of image-text samples, each of which contains one image and two captions.
Compared to previous benchmarks, each wrong caption in SUGARCREPE is manipulated with one of the following manipulations: \textit{SWAP}, \textit{REPLACE} and \textit{ADD}, and debiased to obtain a plausible and fluent text \cite{chatgpt}.
Such a comprehensive benchmark presents a more significant compositional challenge to VLMs.
In experiments, we harnessed Spacy \cite{spacy} to roughly obtain the masks denoting object words, relation words and attribute words.
Results are summarized from Table \ref{sugarcrepe-object} to \ref{sugarcrepe-attribute}.

Generally speaking, given such complex caption pairs, text encoders of VLMs still demonstrated similar behavior on the \textit{REPLACE} manipulation and the \textit{SWAP} manipulation as in previously evaluated benchmarks.  
As for the new \textit{ADD} manipulation, it is expected that adding new object/attribute words should not only change the effect of object/attribute words alone, but also vary the interactions they contributed to.
Results in Table \ref{sugarcrepe-object} and \ref{sugarcrepe-attribute} are consistent with the above intuitions, further demonstrating that text encoders of VLMs were capable of recognizing the compositional differences between captions in a fine-grained manner.

\begin{table}[t]
\caption{Evaluating the compositional knowledge encoded in text encoders of VLMs with the SUGARCREPE dataset (object/relation aspect). 
Here $\rho_{O}$, $\rho_{R}$ and  $\rho_{R\&O}$ represents $\rho (\mathcal{X^T}, \mathcal{Y}^\mathcal{T}_{O})$, $\rho (\mathcal{X^T}, \mathcal{Y}^\mathcal{T}_{R})$ and $\rho (\mathcal{X^T}, \mathcal{Y}^\mathcal{T}_{R\&O})$ for short.
On the object aspect, for the \textit{ADD} manipulation, results show that $Q_O$ and $Q_{R\& O}$ (as well as $\rho_{O}$ and  $\rho_{R\&O}$) had the largest values (shown in bold), showing that text encoders of VLMs recognized both the changes of object words and the interaction changes between object words and relation words, when adding new object words in captions.
As for the \textit{REPLACE} manipulation on the object aspect, $Q_O$ and $Q_{R\& O}$ (as well as $\rho_{O}$ and  $\rho_{R\&O}$) still had the largest values (shown in bold), showing that text encoders of VLMs recognized both the changes of object words and the interaction changes between relation words and object words, when replacing object words in captions.
As for the \textit{SWAP} manipulation on the object aspect, $Q_{R\& O}$ (as well as $\rho_{R\&O}$) had the largest value (shown in bold), showing that text encoders of VLMs recognized the interaction changes between relation words and object words, when swapping object words in captions.
On the relation aspect, for the \textit{REPLACE} manipulation, results show that $Q_R$ and $Q_{R\& O}$ (as well as $\rho_{R}$ and  $\rho_{R\&O}$) had the largest values, showing that when replacing the relation words in captions, text encoders of VLMs successfully recognized both the changes of relation words and the interaction changes between relation words and object words.
The above results show that text encoders of VLMs recognized the compositional differences between complex captions in the SUGARCREPE dataset, similar to our intuitions across different types of manipulations.}
\begin{center}
{\linespread{1.0}
\setlength\tabcolsep{1pt}
\scriptsize
\begin{threeparttable}
\begin{tabular}{c|c | c| c c c |c c c }
\hline
\makecell[c]{Dataset} &Manipulation &{Models}
& $Q_O$ & $Q_R$ & $Q_{R \& O}$
& $\rho_O$ & $\rho_R$ & $\rho_{R\&O}$\\
\hline
\hline
\multirow{15}{*}{\makecell[c]{SUGARCREPE \\ (Object)}}& \multirow{5}{*}{\textit{ADD}} &CLIP  & \textbf{3.3e-2} & 1.1e-2  & \textbf{2.1e-2}& \textbf{0.55}&0.07&\textbf{0.37}  \\
&    & NEGCLIP &\textbf{4.2e-2} & 7.6e-3 & \textbf{1.3e-2}&\textbf{0.66}&0.12& \textbf{0.17} \\
&    & BLIP &\textbf{2.1e-1}&3.0e-2 &\textbf{3.1e-1} &\textbf{0.32}&-0.03& \textbf{0.58}\\
&    & XVLM &\textbf{2.0e-1}&9.9e-2&\textbf{1.8e-1}&\textbf{0.32}&0.07& \textbf{0.23}\\
&    & FLAVA &\textbf{5.5e-1}& 3.0e-1 &\textbf{7.5e-1} &\textbf{0.06}&0.02&\textbf{0.10} \\
\cline{2-9}
& \multirow{5}{*}{\textit{REPLACE}} &CLIP  &\textbf{1.6e-1}   & 9.8e-3  & \textbf{2.9e-2}&\textbf{0.80}&0.09&\textbf{0.37}  \\
&   & NEGCLIP &\textbf{1.5e-1}&7.4e-3 & \textbf{1.7e-2}&\textbf{0.84}&0.11& \textbf{0.32}\\
&    & BLIP &\textbf{8.1e-1}&2.3e-2 &  \textbf{5.7e-1 }&\textbf{0.59}&0.07&\textbf{0.52} \\
&    & XVLM &\textbf{5.8e-1}& 4.7e-2&\textbf{2.3e-1} &\textbf{0.51}&0.03&\textbf{0.24} \\
&    & FLAVA &\textbf{6.8e-1} & 1.3e-1 & \textbf{5.7e-1} &\textbf{0.16}&-0.02& \textbf{0.30}\\
\cline{2-9}
& \multirow{5}{*}{\textit{SWAP}} &CLIP  &9.5e-3 & 9.5e-3 & \textbf{1.3e-2}& 0.29 &0.01&\textbf{0.58}  \\
&   & NEGCLIP &7.1e-3 & 3.5e-3& \textbf{1.7e-2}&0.14&0.01&\textbf{0.51}  \\
&    & BLIP & 4.0e-2 & 1.4e-2&\textbf{1.3e-1} &0.23&-0.05& \textbf{0.65}\\
&    & XVLM &6.6e-2&3.3e-2 &\textbf{1.3e-1}&0.10&0.01&\textbf{0.47} \\
&    & FLAVA &1.3e-1&1.1e-1 &\textbf{3.6e-1} &0.13&0.02&\textbf{0.17} \\
\hline
\multirow{5}{*}{\makecell[c]{SUGARCREPE \\ (Relation)}}& \multirow{5}{*}{\textit{REPLACE}} &CLIP  & 3.9e-3  & \textbf{3.9e-2}   &\textbf{3.0e-2}& 0.03&\textbf{0.47}&\textbf{0.36} \\
&    & NEGCLIP &1.4e-3& \textbf{4.3e-2}& \textbf{2.4e-2}&-0.05&\textbf{0.53}&  \textbf{0.13}\\
&    & BLIP &1.3e-2& \textbf{1.3e-1}& \textbf{3.1e-1}&0.05&\textbf{0.45}& \textbf{0.76}\\
&    & XVLM &3.5e-2& \textbf{1.8e-1}&\textbf{2.2e-1} &0.00&\textbf{0.26}&\textbf{0.44} \\
&    & FLAVA &7.6e-2 & \textbf{4.0e-1}  &\textbf{4.7e-1} &-0.01&\textbf{0.12}& \textbf{0.13}\\
\hline
\end{tabular}
\end{threeparttable}
}
\end{center}
\label{sugarcrepe-object}
\end{table}

\begin{table}[t]
\caption{Evaluating the compositional knowledge encoded in text encoders of VLMs with the SUGARCERPE dataset (attribute aspect). 
Here $\rho_{O}$, $\rho_{A}$ and  $\rho_{A\&O}$ represents $\rho (\mathcal{X^T}, \mathcal{Y}^\mathcal{T}_{O})$, $\rho (\mathcal{X^T}, \mathcal{Y}^\mathcal{T}_{A})$ and $\rho (\mathcal{X^T}, \mathcal{Y}^\mathcal{T}_{A\&O})$ for short.
On the attribute aspect, for the \textit{ADD} manipulation, results show that $Q_A$ and $Q_{A\& O}$ (as well as $\rho_{A}$ and  $\rho_{A\&O}$) had the largest values (shown in bold), showing that text encoders of VLMs recognized both the changes of attribute words and the interaction changes between object words and attribute words, when adding new attribute words in captions.
As for the \textit{REPLACE} manipulation on the attribute aspect, $Q_A$ and $Q_{A\& O}$ (as well as $\rho_{A}$ and  $\rho_{A\&O}$) still had the largest values (shown in bold), showing that text encoders of VLMs recognized both the changes of attribute words and the interaction changes between attribute words and object words, when replacing attribute words in captions.
As for the \textit{SWAP} manipulation on the attribute aspect, $Q_{A\& O}$ (as well as $\rho_{A\&O}$) had the largest value (shown in bold), showing that text encoders of VLMs recognized the interaction changes between attribute words and object words, when swapping attribute words in captions.
The above results show that text encoders of VLMs also correctly recognized the attribute-wise compositional difference within complex caption pairs in the SUGARCERPE dataset.}
\begin{center}
{\linespread{1.0}
\setlength\tabcolsep{1pt}
\scriptsize
\begin{threeparttable}
\begin{tabular}{c|c | c| c c c |c c c }
\hline
\makecell[c]{Dataset)} &Manipulation &{Models}
& $Q_O$ & $Q_A$ & $Q_{A \& O}$
& $\rho_O$ & $\rho_A$ & $\rho_{A\&O}$\\
\hline
\hline
\multirow{15}{*}{\makecell[c]{SUGARCREPE \\ (Attribute)}}& \multirow{5}{*}{\textit{ADD}} &CLIP  & 2.8e-3  & \textbf{2.4e-2} &\textbf{2.5e-2}& 0.05& \textbf{0.38}&\textbf{0.34} \\
&    & NEGCLIP &1.8e-3&\textbf{3.5e-2} & \textbf{1.8e-2}&0.00&\textbf{0.46}& \textbf{0.14}\\
&    & BLIP &2.7e-2&\textbf{1.1e-1} &\textbf{2.8e-1} &-0.03&\textbf{0.49}& \textbf{0.66}\\
&    & XVLM &5.1e-2&\textbf{1.3e-1} & \textbf{2.1e-1} &0.00&\textbf{0.22}& \textbf{0.35}\\
&    & FLAVA &2.2e-1 & \textbf{3.3e-1} &\textbf{5.3e-1} &-0.03&\textbf{0.13}&\textbf{0.07} \\
\cline{2-9}
& \multirow{5}{*}{\textit{REPLACE}} &CLIP  & 3.3e-3  & \textbf{6.1e-2}   &\textbf{2.4e-2} & 0.10 & \textbf{0.67} & \textbf{0.36} \\
&   & NEGCLIP &2.9e-3&\textbf{8.0e-2} &\textbf{1.8e-2} &0.09&\textbf{0.72}&  \textbf{0.16}\\
&    & BLIP &5.6e-3 & \textbf{3.8e-1} & \textbf{5.0e-1} &-0.03& \textbf{0.50}& \textbf{0.71} \\
&    & XVLM &1.3e-2&\textbf{2.8e-1} & \textbf{2.6e-1}&-0.06&\textbf{0.31}&\textbf{0.47} \\
&    & FLAVA &4.8e-2 & \textbf{4.7e-1} & \textbf{4.5e-1}  &-0.02&\textbf{0.29}&\textbf{0.20} \\
\cline{2-9}
& \multirow{5}{*}{\textit{SWAP}} &CLIP  & 1.7e-2 &  1.7e-2 &\textbf{2.2e-2}& 0.13&  0.17&\textbf{0.47}\\
&   & NEGCLIP &9.8e-3 &1.3e-2 &\textbf{2.0e-2}  &0.20&0.09&\textbf{0.43}  \\
&    & BLIP &4.5e-2&7.5e-2 & \textbf{7.7e-1}&-0.01&0.23& \textbf{0.82}\\
&    & XVLM &8.8e-2&1.3e-1 & \textbf{4.3e-1}&-0.01&0.16&\textbf{0.56} \\
&    & FLAVA &9.8e-2 & 2.0e-1 &\textbf{4.3e-1} &-0.04&0.06& \textbf{0.18}\\
\hline
\end{tabular}
\end{threeparttable}
}
\end{center}
\label{sugarcrepe-attribute}
\end{table}

\begin{table}[!h]
\caption{Evaluating the mutual compositional knowledge encoded in image encoders and text encoders of VLMs with the Visual Genome Relation dataset. The
maximum metric values are shown in bold.}
\begin{center}
{\linespread{1.0}
\setlength\tabcolsep{2pt}
\scriptsize
\begin{threeparttable}
\begin{tabular}{c | c| c c c }
\hline
{Dataset} &{Models}
& $Q_{\mathcal{T}:R\&O \xrightarrow{} \mathcal{I}:O_1}$ & $Q_{\mathcal{T}:R\&O \xrightarrow{} \mathcal{I}:O_2}$ & $Q_{\mathcal{T}:R\&O \xrightarrow{} \mathcal{I}:O_1\&O_2}$
\\
\hline
\hline
\multirow{5}{*}{Visual Genome Relation}&CLIP   &\textbf{3.7e-1}&3.3e-1 &1.6e-1 \\
&     NEGCLIP &\textbf{7.3e-1}&5.9e-1 &2.2e-1\\
&     BLIP &\textbf{1.8e-1}&1.7e-1 &8.5e-2\\
&     XVLM &\textbf{3.4e-1}&2.9e-1 &1.5e-1  \\
&     FLAVA &\textbf{3.2e-1}&3.1e-1 &2.7e-1\\
\hline
\end{tabular}
\end{threeparttable}
}
\end{center}
\label{arosugarcrepe-relation-t2i}
\end{table}

\section{More results on evaluating the mutual compositional knowledge between text encoders and image encoders}
In this section, we provide more results to further evaluate the mutual compositional knowledge between text encoders and image encoders.
To this end, we exploited the Visual Genome Relation dataset to conduct further analyses on a larger scale.
Since each sample in the dataset contains one image and two captions individually, we mainly used ${Q_{\mathcal{T}:R\&O \xrightarrow{}\mathcal{I}:(\cdot)}}$ to evaluate how image encoders considered the compositional knowledge of text encoders.
For the Visual Genome Relation dataset, we obtained the segmentation mask for each object with the help of SAM \cite{SAM}, similar to our annotated EQBEN dataset.
The newly annotated dataset contained 2000 samples in total.
The newly added segmentation masks are visualized in Figure \ref{fig:cc}.
In this way, as shown in Table \ref{arosugarcrepe-relation-t2i}, ${Q_{\mathcal{T}:R\&O \xrightarrow{}\mathcal{I}:O_1\&O_2}}$ failed to have the largest value than ${Q_{\mathcal{T}:R\&O \xrightarrow{}\mathcal{I}:O_1}}$ and ${Q_{\mathcal{T}:R\&O \xrightarrow{}\mathcal{I}:O_2}}$, showing that in terms of object relations, image encoders did not have the corresponding compositional knowledge of text encoders. 
Instead, image encoders tended to consider the interaction between object words and relation words learned by text encoders as the object representations on images (\emph{i.e.}, ${Q_{\mathcal{T}:R\&O \xrightarrow{}\mathcal{I}:O_1}}$ being the largest among metrics).
The above results were consistent with the results in Table \ref{eqben-relation-t2i} in terms of ${Q_{\mathcal{T}:R\&O \xrightarrow{}\mathcal{I}:(\cdot)}}$.

\section{More results on recent VLMs}
In this section, we further conducted experiments on two more recent VLMs: FLIP \cite{li2023scaling} and LaCLIP \cite{fan2023improving}. 
Specifically, we followed our proposed methods from Section 4 to 6 with the same evaluation benchmarks, comprehensively evaluating the compositional knowledge of text encoders, the compositional knowledge of image encoders and also, whether text encoders and image encoders have mutually matching compositional knowledge. We summarized the results in Table \ref{tab:recent_vlms_text_encoders}, Table \ref{tab:recent_vlms_image_encoders} and Table \ref{tab:recent_vlms_text_image_encoders} respectively.
Experimental results are consistent with our findings in the main paper. Specifically,
(1) text encoders of FLIP and LaCLIP showed excellent compositional reasoning capabilities, able to recognize the dominant compositional differences between input texts like human understanding;
(2) image encoders of FLIP and LaCLIP demonstrated weaker compositional reasoning capabilities;
(3) image encoders and text encoders of FLIP and LaCLIP did not exhibit mutually matching compositional knowledge.
The above experiments further strengthen the reliability of the findings in the main paper.

\begin{table}[h]
\caption{Evaluating the compositional knowledge of text encoders of recent VLMs.}
\begin{center}
{\linespread{1.0}
\setlength\tabcolsep{6pt}
\scriptsize
\begin{threeparttable}
\begin{tabular}{l l c c c}
\hline
{Dataset} &{Models}
& $Q_O$ & $Q_R$ & $Q_{R \& O}$ \\
\hline
\hline
\multirow{2}{*}{Visual Genome Relation}&FLIP&	1.6e-2	&9.9e-5	&\textbf{4.2e-2}  \\
&LaCLIP&3.7e-3&	2.5e-6&\textbf{8.3e-3} \\
\hline
\end{tabular}
\end{threeparttable}
}
\end{center}
\label{tab:recent_vlms_text_encoders}
\end{table}

\begin{table}[h]
\caption{Evaluating the compositional knowledge of image encoders of recent VLMs.}
\begin{center}
{\linespread{1.0}
\setlength\tabcolsep{6pt}
\scriptsize
\begin{threeparttable}
\begin{tabular}{l l c c c}
\hline
{Dataset} &{Models}
& $D_{O_1}$ & $D_{O_2}$ & $D_{O_1 \& O_2}$ \\
\hline
\hline
\multirow{2}{*}{EQBEN}&FLIP	&3.4e-2&	\textbf{5.2e-2}&	3.3e-2  \\
&LaCLIP&1.5e-2&	\textbf{2.2e-2}&	2.1e-2 \\
\hline
\end{tabular}
\end{threeparttable}
}
\end{center}
\label{tab:recent_vlms_image_encoders}
\end{table}

\begin{table}[h]
\caption{Evaluating whether image encoders and text encoders of recent VLMs possess mutually matching compositional knowledge.}
\begin{center}
{\linespread{1.0}
\setlength\tabcolsep{0.5pt}
\scriptsize
\begin{threeparttable}
\begin{tabular}{c | c| c c c  }
\hline
{Dataset} &{Models}
& $Q_{\mathcal{T}:R\&O \xrightarrow{} \mathcal{I}:O_1}$ & $Q_{\mathcal{T}:R\&O \xrightarrow{} \mathcal{I}:O_2}$ & $Q_{\mathcal{T}:R\&O \xrightarrow{} \mathcal{I}:O_1\&O_2}$ \\
\hline
\hline
\multirow{2}{*}{EQBEN}&FLIP&   2.2e-1&	\textbf{5.5e-1}&	4.0e-1 \\
&     LaCLIP &5.8e-1&	\textbf{2.0e0}&	7.4e-1  \\
\hline
\hline
{Dataset} &{Models}
&  $D_{\mathcal{I}:O_1\&O_2 \xrightarrow{} \mathcal{T}:R}$ & $D_{\mathcal{I}:O_1\&O_2 \xrightarrow{} \mathcal{T}:O}$ & $D_{\mathcal{I}:O_1\&O_2 \xrightarrow{} \mathcal{T}:R\&O}$\\
\hline
\hline
\multirow{2}{*}{EQBEN}&FLIP  &8.1e-1&	\textbf{2.1e0}&	1.0e0 \\
&     LaCLIP  &7.8e-1&	\textbf{4.0e0}&	8.8e-1\\
\hline
\end{tabular}
\end{threeparttable}
}
\end{center}
\label{tab:recent_vlms_text_image_encoders}
\end{table}

\section{Experimental details}
In the main paper, we evaluated the following state-of-the-art Vision Language Models (VLMs): CLIP\footnote{https://github.com/openai/CLIP/} \cite{clip}, NEGCLIP\footnote{https://github.com/mertyg/vision-language-models-are-bows} \cite{aro}, BLIP\footnote{https://github.com/salesforce/BLIP} \cite{blip}, XVLM\footnote{https://github.com/zengyan-97/X-VLM/} \cite{xvlm}, FLAVA\footnote{https://github.com/apsdehal/flava-tutorials/blob/main/winoground-flava-example.ipynb} \cite{flava}, all of which were obtained from their officially released checkpoints.
For reference, we also provide the compositional performance of these VLMs with the benchmarks exploited in the main paper.
Specifically, we used $ACC_\mathcal{T}$ to measure the accuracy of VLMs picking up the correct caption when given one image with two captions.
We also used $ACC_\mathcal{I}$ to measure the accuracy of VLMs picking up the correct image when given one caption with two images.
Results in Tab. \ref{aro-performance}-\ref{vlchecklist-performance-or} show that these state-of-the-art VLMs performed poorly on these benchmarks, exposing their weakness in understanding the compositional information of input variables.

\begin{table}[!h]
\caption{Evaluating the compositional performance of VLMs with the Visual Genome Relation dataset and Visual Genome Attribution dataset. }
\begin{center}
{\linespread{1.0}
\setlength\tabcolsep{4pt}
\scriptsize
\begin{threeparttable}
\begin{tabular}{l l c c c c}
\hline
{Dataset} &{Models}
& $ACC_\mathcal{T}$ & {Dataset} &{Models}& $ACC_\mathcal{T}$ \\
\hline
\hline
\multirow{5}{*}{Visual Genome Relation}&CLIP  &50.65\% &\multirow{5}{*}{Visual Genome Attribution}&CLIP  &60.98\%   \\
&NEGCLIP &58.33\% &  & NEGCLIP &70.87\%\\
&BLIP &55.30\% &  & BLIP &87.12\%\\
&XVLM &58.65\% &  &XVLM &89.51\% \\
&FLAVA&43.79\% &  &FLAVA &68.84\% \\
\hline
\end{tabular}
\end{threeparttable}
}
\end{center}
\label{aro-performance}
\end{table}

\begin{table*}[!h]
\caption{Evaluating the compositional performance of VLMs with the VL-CheckList dataset (attribute aspect). }
\begin{center}
{\linespread{1.0}
\setlength\tabcolsep{2pt}
\scriptsize
\begin{threeparttable}
\begin{tabular}{l l l c l l c l l c l l c l l c}
\hline
{Dataset} &{Category}&{Models}
& $ACC_\mathcal{T}$ & {Category} &{Models}& $ACC_\mathcal{T}$& {Category} &{Models}& $ACC_\mathcal{T}$& {Category} &{Models}& $ACC_\mathcal{T}$& {Category} &{Models}& $ACC_\mathcal{T}$ \\
\hline
\hline
\multirow{5}{*}{\makecell[c]{VL-CheckList \\ (Attribute)}}&\multirow{5}{*}{Color} &CLIP  &71.20\% &\multirow{5}{*}{Material}&CLIP  &66.32\% &\multirow{5}{*}{Size}&CLIP  &60.85\%&\multirow{5}{*}{State}&CLIP  &61.68\%&\multirow{5}{*}{Action}&CLIP  &73.46\%  \\
&&NEGCLIP &73.59\% &  & NEGCLIP &75.87\%&& NEGCLIP &66.06\%&& NEGCLIP &70.60\%&& NEGCLIP &75.34\%\\
&&BLIP &86.49\% &   &BLIP &89.85\%&&BLIP &72.13\%&&BLIP &78.20\%&&BLIP &80.98\%\\
&&XVLM &86.41\% &  &XVLM &87.54\%&&XVLM &76.67\%&&XVLM &77.77\%&&XVLM &82.44\% \\
&&FLAVA&69.76\% &  &FLAVA &56.05\%&&FLAVA &37.77\%&&FLAVA &50.05\%&&FLAVA &57.37\% \\
\hline
\end{tabular}
\end{threeparttable}
}
\end{center}
\label{vlchecklist-performance-a}
\end{table*}

\begin{table*}[h]
\caption{Evaluating the compositional performance of VLMs with the SUGARCREPE dataset (object/relation aspect). }
\begin{center}
{\linespread{1.0}
\setlength\tabcolsep{1pt}
\scriptsize
\begin{threeparttable}
\begin{tabular}{c c l c c l c c l c| c c l c}
\hline
{Dataset} &{Manipulation}&{Models}
& $ACC_\mathcal{T}$ & {Manipulation} &{Models}& $ACC_\mathcal{T}$& {Manipulation} &{Models}& $ACC_\mathcal{T}$& {Dataset} & {Manipulation} &{Models}& $ACC_\mathcal{T}$ \\
\hline
\hline
\multirow{5}{*}{\makecell[c]{SUGARCREPE \\ (Object)}}&\multirow{5}{*}{\textit{SWAP}} &CLIP  &62.70\% &\multirow{5}{*}{\textit{ADD}}&CLIP  &76.88\% &\multirow{5}{*}{\textit{REPLACE}}&CLIP  &91.85\%&\multirow{5}{*}{\makecell[c]{SUGARCREPE \\ (Relation)}}&\multirow{5}{*}{\textit{REPLACE}}&CLIP  &69.18\%  \\
&&NEGCLIP &77.05\% &  & NEGCLIP &88.12\%&& NEGCLIP &92.70\%&&& NEGCLIP &74.71\%\\
&&BLIP &82.04\% &   &BLIP &95.98\%&&BLIP &98.16\%&&&BLIP &86.32\%\\
&&XVLM &86.12\% &  &XVLM &95.49\%&&XVLM &97.73\%&&&XVLM &85.89\% \\
&&FLAVA&70.20\% &  &FLAVA &90.64\%&&FLAVA &90.74\%&&&FLAVA &64.07\% \\
\hline
\end{tabular}
\end{threeparttable}
}
\end{center}
\label{sugarcrepe-performance-or}
\end{table*}

\begin{table*}[!h]
\caption{Evaluating the compositional performance of VLMs with the SUGARCREPE dataset (attribute aspect) and EQBEN dataset (relation aspect). 
The performance on the EQBEN dataset was evaluated on our newly annotated subset (several samples are shown in Figure \ref{fig:bb}).}
\begin{center}
{\linespread{1.0}
\setlength\tabcolsep{2pt}
\scriptsize
\begin{threeparttable}
\begin{tabular}{c c l c c l c c l c| c l c c}
\hline
{Dataset} &{Manipulation}&{Models}
& $ACC_\mathcal{T}$ & {Manipulation} &{Models}& $ACC_\mathcal{T}$& {Manipulation} &{Models}& $ACC_\mathcal{T}$& {Dataset} & {Manipulation} &{$ACC_\mathcal{I}$}& $ACC_\mathcal{T}$ \\
\hline
\hline
\multirow{5}{*}{\makecell[c]{SUGARCREPE \\ (Attribute)}}&\multirow{5}{*}{\textit{SWAP}} &CLIP  &64.94\% &\multirow{5}{*}{\textit{ADD}}&CLIP  &67.60\% &\multirow{5}{*}{\textit{REPLACE}}&CLIP  &81.94\%&\multirow{5}{*}{\makecell[c]{EQBEN \\ (Relation)}}&CLIP  &53.85\% &55.90\%  \\
&&NEGCLIP &76.07\% &  & NEGCLIP &81.44\%&& NEGCLIP &85.21\%&& NEGCLIP &53.85\%&62.15\%\\
&&BLIP &94.33\% &   &BLIP &90.18\%&&BLIP &93.05\%&&BLIP &59.72\%&60.76\%\\
&&XVLM &93.10\% &  &XVLM &86.76\%&&XVLM &91.35\%&&XVLM &61.46\% &66.67\%\\
&&FLAVA&81.90\% &  &FLAVA &59.08\%&&FLAVA &76.15\%&&FLAVA &54.86\%&61.46\% \\
\hline
\end{tabular}
\end{threeparttable}
}
\end{center}
\label{sugarcrepe-performance-a_eqben}
\end{table*}

\begin{table}[!h]
\caption{Evaluating the compositional performance of VLMs with the VL-CheckList dataset (object/relation aspect). }
\begin{center}
{\linespread{1.0}
\setlength\tabcolsep{3pt}
\scriptsize
\begin{threeparttable}
\begin{tabular}{l l c c c c c c}
\hline
{Dataset} &{Category}&{Models}
& $ACC_\mathcal{T}$ & {Category} &{Models}& $ACC_\mathcal{T}$ \\
\hline
\hline
\multirow{5}{*}{\makecell[c]{VL-CheckList \\ (Object)}}&\multirow{5}{*}{Location} &CLIP  &88.80\% &\multirow{5}{*}{Size}&CLIP  &89.02\%   \\
&&NEGCLIP &90.72\% &  & NEGCLIP &89.50\%\\
&&BLIP &92.52\% &   &BLIP &92.25\%\\
&&XVLM &92.85\% &  &XVLM &92.08\% \\
&&FLAVA&73.21\% &  &FLAVA &71.34\% \\
\hline
\hline
\multirow{5}{*}{\makecell[c]{VL-CheckList \\ (Relation)}}&\multirow{5}{*}{Action} &CLIP  &77.05\% &\multirow{5}{*}{Spatial}&CLIP  &55.77\%   \\
&&NEGCLIP &81.57\% &  & NEGCLIP &60.80\%\\
&&BLIP &81.19\% &   &BLIP &61.68\%\\
&&XVLM &77.56\% &  &XVLM &74.90\% \\
&&FLAVA&33.55\% &  &FLAVA &58.45\% \\
\hline
\end{tabular}
\end{threeparttable}
}
\end{center}
\label{vlchecklist-performance-or}
\end{table}

\section{Future studies}
In this paper, we have obtained and validated several insights to explain the poor compositional reasoning capabilities of VLMs, which we believe could provide beneficial guidance for future studies.
However, considering the large gap between theory and practice, significant research efforts are still required to achieve the breakthrough in practice based on our analyses, which we plan to leave to future endeavors. To this end, we introduce two promising solutions as follows: (1) improving image encoders' sensitivities to compositionality changes, instead of text encoders. 
To this end, one possible solution is to design modules to specifically approximate the Harsanyi dividends of different visual patterns in image encoders, drawing inspiration from \cite{chen2023harsanyinet}. This approach can allow us to explicitly extract the casual effects of different visual patterns during the training phase, making the improvements of visual compositionality sensitivities more accessible;
(2) enhancing the alignment of compositional knowledge between text encoders and image encoders of VLMs. To this end, one possible solution is to leverage our metrics in Eq. \ref{eq:Q}- Eq. \ref{eq-di2t} as an auxiliary training loss for VLMs. However, implementing this approach requires a large scale of training data featuring subtle changes of compositionality with detailed textual and visual annotations, which underscores the need for additional research efforts in this direction.

%%%%%%%%%%%%%%%%%%%%%%%%%%%%%%%%%%%%%%%%%%%%%%%%%%%%%%%%%%%%%%%%%%%%%%%%%%%%%%%
%%%%%%%%%%%%%%%%%%%%%%%%%%%%%%%%%%%%%%%%%%%%%%%%%%%%%%%%%%%%%%%%%%%%%%%%%%%%%%%

\end{document}